\documentclass{article}

\usepackage{times}
\usepackage{graphicx} 
\usepackage{subfigure} 
\usepackage{amssymb}
\usepackage{amsmath}
\usepackage{amsthm}
\usepackage{framed}
\usepackage{adjustbox}

\usepackage{natbib}

\usepackage{algorithm}
\usepackage{algorithmic}

\usepackage{hyperref}

\newcommand{\e}[1]{{\mathbb E}\left[ #1 \right]}

\usepackage[accepted]{icml2017}

\icmltitlerunning{Sub-sampled Cubic Regularization for Non-convex Optimization}

\newcommand{\norm}[1]{{}\left\| #1 \right\|}

\newcommand{\g}{{\bf g}}

\newcommand{\s}{{\bf s}}

\newcommand{\w}{{\bf w}}

\newcommand{\x}{{\bf x}}
\newcommand{\y}{{\bf y}}
\newcommand{\z}{{\bf z}}
\newcommand{\eb}{{\bf e}}
\newcommand{\evec}{{\bf e}}
\newcommand{\supervector}{{\bf v}}

\newcommand{\vb}{{\bf v}}

\def\Bm{{\bf B}}
\def\Hm{{\bf H}}
\def\Qm{{\bf Q}}
\def\Am{{\bf A}}
\def\Zm{{\bf Z}}
\def\Ym{{\bf Y}}
\def\Xm{{\bf X}}

\def\Im{{\bf I}}

\newcommand{\R}{{\mathbb{R}}}

\newtheorem{theorem}{Theorem}
\newtheorem{lemma}[theorem]{Lemma}

\newtheorem{assumption}[theorem]{Assumption}

\graphicspath{{./figures/}}

\begin{document} 

\twocolumn[
\icmltitle{Sub-sampled Cubic Regularization for Non-convex Optimization}



\icmlsetsymbol{equal}{*}

\begin{icmlauthorlist}
\icmlauthor{Jonas Moritz Kohler}{to}
\icmlauthor{Aurelien Lucchi}{to}
\end{icmlauthorlist}

\icmlaffiliation{to}{Department of Computer Science, ETH Zurich, Switzerland}

\icmlcorrespondingauthor{Jonas Moritz Kohler}{jonas.kohler@student.kit.edu}
\icmlcorrespondingauthor{Aurelien Lucchi}{aurelien.lucchi@inf.ethz.ch}

\icmlkeywords{Non-convex optimization, stochastic, trust region, cubic regularization}

\vskip 0.3in
]



\printAffiliationsAndNotice{}  

\newcommand{\methodname}{{\textsc{SCR }}}

\begin{abstract}
We consider the minimization of non-convex functions that typically arise in machine learning. Specifically, we focus our attention on a variant of trust region methods known as cubic regularization. This approach is particularly attractive because it escapes~\textit{strict} saddle points and it provides stronger convergence guarantees than first- and second-order as well as classical trust region methods. However, it suffers from a high computational complexity that makes it impractical for large-scale learning. Here, we propose a novel method that uses sub-sampling to lower this computational cost. By the use of concentration inequalities we provide a sampling scheme that gives sufficiently accurate gradient and Hessian approximations to retain the strong global and local convergence guarantees of cubically regularized methods. To the best of our knowledge this is the first work that gives global convergence guarantees for a sub-sampled variant of cubic regularization on non-convex functions. Furthermore, we provide experimental results supporting our theory.
\end{abstract} 


\section{Introduction}

In this paper we address the problem of minimizing an objective function of the form
\begin{equation}
\x^* = \arg \min_{\x \in \R^d}  \left [ f(\x) := \frac1n \sum_{i=1}^n f_i(\x) \right ],
\label{eq:f_x}
\end{equation}
where $f(\x) \in C^2(\R^d, \R)$ is a not necessarily convex, (regularized) loss function over $n$ datapoints.
Stochastic Gradient Descent (SGD) is a popular method to optimize this type of objective especially in the context of large-scale learning when $n$ is very large. Its convergence properties are well understood for convex functions, which arise in many machine learning applications \cite{nesterov2004introductory}. However, non-convex functions are also ubiquitous and have recently drawn a lot of interest due to the growing success of deep neural networks. Yet, non-convex functions are extremely hard to optimize due to the presence of saddle points and local minima which are not global optima~\cite{dauphin2014identifying, choromanska2015loss}. In fact, the work of~\cite{hillar2013most} showed that even a degree four polynomial can be NP-hard to optimize. Instead of aiming for a global minimizer, we will thus seek for a local optimum of the objective. In this regard, a lot of attention has focused on a specific type of functions known as strict saddle functions or ridable functions~\cite{ge2015escaping, sun2015nonconvex}. These functions are characterized by the fact that the Hessian of every saddle point has a negative eigenvalue. Geometrically this means that there is a direction of negative curvature where decreasing function values can be found. Examples of strict saddle functions include dictionary learning, orthogonal tensor decomposition and generalized phase retrieval ~\cite{ge2015escaping, sun2015nonconvex}.

In this work, we focus our attention on trust region methods to optimize Eq.~\ref{eq:f_x}. These methods construct and optimize a local model of the objective function within a region whose radius depends on how well the model approximates the real objective. One of the keys for efficiency of these methods is to pick a model that is comparably easy to optimize, such as a quadratic function~\cite{conn2000trust}.
Following the trust region paradigm, cubic regularization methods~\cite{nesterov2006cubic, cartis2011adaptive} suggest finding the step $\s_k$ that minimizes a cubic model  of the form
\begin{equation}
m_k(\s_k) := f(\x_k) + \s_k^\intercal \nabla f(\x_k) + \frac{1}{2} \s_k^\intercal \Hm_k \s_k +\frac{\sigma_k}{3} \norm{\s_k}^3,
\label{eq:cubic_model}
\end{equation}
where $\Hm_k := \nabla^2 f(\x_k)$ and $\sigma_k > 0$~\footnote{In the work of~\cite{nesterov2006cubic}, $\sigma_k$ is assumed to be the Lipschitz constant of the Hessian in which case the model defined in Eq.~\ref{eq:cubic_model} is a global overestimator of the objective, i.e. $f(\x) \leq m(\x) \; \forall \x \in \mathbb{R}^d$. We will elaborate on the role of $\sigma_k$ in~\cite{cartis2011adaptive} later on.}.

\cite{nesterov2006cubic} were able to show that, if the step is computed by globally minimizing the cubic model and if the Hessian $\Hm_k$ is globally Lipschitz continuous, Cubic regularization methods possess the best known worst case complexity to solve Eq.~\ref{eq:f_x}: an overall worst-case iteration count of order $\epsilon^{-3/2}$ for generating $\| \nabla f(\x_k) \| \leq \epsilon$, and of order $\epsilon^{-3}$ for achieving approximate nonnegative curvature.
However, minimizing Eq.~\ref{eq:cubic_model} in an exact manner impedes the performance of this method for machine learning applications as it requires access to the full Hessian matrix. More recently, \cite{cartis2011adaptive} presented a method (hereafter referred to as ARC) which relaxed this requirement by assuming that one can construct an approximate Hessian $\Bm_k$ that is sufficiently close to $\Hm_k$ in the following way:
\begin{equation}
\norm{(\Bm_k-\Hm_k) \s_k} \leq C \norm{\s_k}^2, \:\forall k\geq 0, C>0
\label{eq:Strong_agreement_g_0}
\end{equation}
Furthermore, they showed that it is sufficient to find an approximate minimizer by applying a Lanczos method to build up evolving Krylov spaces, which can be constructed in a Hessian-free manner, i.e. by accessing the Hessians only indirectly via matrix-vector products.
However there are still two obstacles for the application of ARC in the field of machine learning: (1) The cost of the Lanczos process increases linearly in $n$ and is thus not suitable for large datasets and (2) there is no theoretical guarantee that quasi-Newton approaches satisfy Eq.~\ref{eq:Strong_agreement_g_0} and~\cite{cartis2011adaptive} do not provide any alternative approximation technique.

In this work, we make explicit use of the finite-sum structure of Eq. \ref{eq:f_x} by applying a sub-sampling technique in order to provide guarantees for machine learning applications. Towards this goal, we make the following contributions:
\begin{itemize}
\itemsep0em
\item We provide a theoretical Hessian sampling scheme that is guaranteed to satisfy Eq.~\ref{eq:Strong_agreement_g_0} with high probability.
\item We extend the analysis to inexact gradients and prove that the convergence guarantees of~\cite{nesterov2006cubic, cartis2011adaptive} can be retained.
\item Since the dominant iteration cost lie in the construction of the Lanczos process and increase linearly in $n$, we lower the computational cost significantly by reducing the number of samples used in each iteration.
\item Finally, we provide experimental results demonstrating significant speed-ups compared to standard first and second-order optimization methods for various convex and non-convex objectives.
\end{itemize}


\section{Related work}
\label{sec:related_work}

\paragraph{Sampling techniques for first-order methods.}
In large-scale learning, when $n \gg d$ most of the computational cost of traditional deterministic optimization methods is spent in computing the exact gradient information. A common technique to address this issue is to use sub-sampling in order to compute an unbiased estimate of the gradient. The simplest instance is Stochastic Gradient Descent (SGD) whose convergence does not depend on the number of datapoints $n$. However, the variance in the stochastic gradient estimates slows its convergence down. The work of~\cite{friedlander2012hybrid} explored a sub-sampling technique for gradient descent in the case of convex functions, showing that it is possible to maintain the same convergence rate as full-gradient descent by carefully increasing the sample size over time. Another way to recover a linear rate of convergence for strongly-convex functions is to use variance-reduced methods~\cite{johnson2013, defazio2014saga, roux2012stochastic, hofmann2015variance, daneshmand2016small}. Recently, the convergence of SGD and its variance-reduced counterparts has also been extended to non-convex functions~\cite{ghadimi2013stochastic, reddi2016stochastic} but the techniques used in these papers require using a randomized sampling scheme which is different from what is typically used in practice. Furthermore, the guarantees these methods provide are only in terms of convergence to critical points. However, the work of~\cite{ge2015escaping, sun2015nonconvex} recently showed that SGD can achieve stronger guarantees in the case of strict saddle functions. Yet, the convergence rate has a polynomial dependency to the dimension $d$ and the smallest eigenvalue of the Hessian which can make this method fairly impractical.

\paragraph{Second-order methods.}
For second-order methods, the problem of avoiding saddle points is even worse as they might be attracted by saddle points or even points of local maximizers~\cite{dauphin2014identifying}. Another predominant issue is the computation (and perhaps storage) of the Hessian matrix, which requires $O(nd^2)$ operations as well as computing the inverse of the Hessian, which requires $O(d^3)$ computations. Quasi-Newton methods such as the well-known (L-)BFGS algorithm partially address this issue by requiring $O(nd + d^2)$ per-iteration cost~\cite{nesterov2004introductory} instead of $O(nd^2 + d^3)$. An increasingly popular alternative is to use sub-sampling techniques to approximate the Hessian matrix, such as done for example in~\cite{byrd2011use} and \cite{erdogdu2015convergence}. The latter method, named NewSamp, approximates the Hessian with a low-rank approximation which reduces the complexity per iteration to $O(nd + |S|d^2)$ with $|S|$ being the sample size~\footnote{Note that this method still requires $O(nd)$ computation for the gradient as it only subsamples the Hessian.}. Although this is a significant reduction in terms of complexity, NewSamp yields a composite convergence rate: quadratic at first but only linear near the minimizer. Unlike NewSamp, our sampling scheme yields a locally quadratic rate of convergence (as well as faster global convergence). Our analysis also does not require using exact gradients and can thus further reduce the complexity per iteration.

\paragraph{Cubic regularization and trust region methods.}
Trust region methods are among the most effective algorithmic frameworks to avoid pitfalls such as local saddle points in non-convex optimization. Classical versions iteratively construct a local quadratic model and minimize it within a certain radius wherein the model is trusted to be sufficiently similar to the actual objective function. This is equivalent to minimizing the model function with a suitable \textit{quadratic} penalty term on the stepsize. Thus, a natural extension is the cubic regularization method introduced by~\cite{nesterov2006cubic} that uses a \textit{cubic} over-estimator of the objective function as a regularization technique for the computation of a step to minimize the objective function. The drawback of their method is that it requires computing the exact minimizer of Eq.~\ref{eq:cubic_model}, thus requiring the exact gradient and Hessian matrix. However finding a global minimizer of the cubic model $m_k(\s)$ may not be essential in practice and doing so might be prohibitively expensive from a computational point of view. ~\cite{cartis2011adaptive} introduced a method named ARC which relaxed this requirement by letting $\s_k$ be an approximation to the minimizer. The model defined by the adaptive cubic regularization method introduced two further changes. First, instead of computing the exact Hessian $\Hm_k$ it allows for a symmetric approximation $\Bm_k$. Second, it introduces a dynamic positive parameter $\sigma_k$ instead of using the global Lipschitz constant $L$.

There have been efforts to further reduce the computational complexity of this problem. For example, \cite{agarwal2016finding} refined the approach of~\cite{nesterov2006cubic} to return an approximate local minimum in time which is linear in the input representation. Similar improvements have been made by ~\cite{carmon2016gradient} and ~\cite{hazan2016linear}. These methods provide alternatives to minimize the cubic model and can thus be seen as complementary to our approach. Finally, the work of~\cite{blanchet2016convergence} proposed a stochastic variant of a trust region method but their analysis does not specify any accuracy level required for the estimation of the stochastic Hessian. ~\cite{cartis2017global} also analyzed a probabilistic cubic regularization variant that allows approximate second-order models but they did not provide an explicit derivation of sampling conditions.


\section{Formulation}
\label{sec:formulation}

We are interested in optimizing Eq.~\ref{eq:f_x} in a large-scale setting when the number of datapoints $n$ is very large such that the cost of solving Eq.~\ref{eq:cubic_model} exactly becomes prohibitive. In this regard we identify a sampling scheme that allows us to retain the convergence results of deterministic trust region and cubic regularization methods, including quadratic local convergence rates and global second-order convergence guarantees as well as worst-case complexity bounds. A detailed theoretical analysis is given in Section ~\ref{sec:analysis}. Here we shall first state the algorithm itself and elaborate further on the type of local nonlinear models we employ as well as how these can be solved efficiently. 

\subsection{Objective function}

Instead of using deterministic gradient and Hessian information as in Eq.~\ref{eq:cubic_model}, we use unbiased estimates of the gradient and Hessian constructed from two independent sets of points denoted by $S_g$ and $S_B$. We then construct a local cubic model that is (approximately) minimized in each iteration:
\begin{equation} \label{eq:subproblem}
m_k(\s_k) := f(\x_k) + \s_k^\intercal \g_k + \frac{1}{2} \s_k^\intercal \Bm_k \s_k + \frac{\sigma_k}{3}  \norm{\s_k}^3
\end{equation}
where $\g_k := \frac{1}{|S_g|} \sum_{i \in S_g} \nabla f_i(\x_k)$ \vspace{2mm} \\ \;\; and \;\; $\Bm_k := \frac{1}{|S_B|} \sum_{i \in S_B} \nabla^2 f_i(\x_k)$.

The model derivative with respect to $\s_k$ is defined as:
\begin{equation}\label{eq.subproblem_abl}
\nabla m_k(\s_k) = \g_k + \Bm_k \s_k + \lambda \s_k \text{  ,where } \lambda=\sigma_k\norm{\s_k}.
\end{equation}

\subsection{Algorithm}

Our Sub-sampled Cubic Regularization approach (SCR) is presented in Algorithm~\ref{alg:src}. At iteration step $k$, we sub-sample two sets of datapoints from which we compute a stochastic estimate of the gradient and the Hessian. We then solve the problem in Eq.~\ref{eq:subproblem} \textit{approximately} using the method described in Section ~\ref{sec:approximate_minimization} and update the regularization parameter $\sigma_k$ depending on how well the model approximates the real objective. In particular, very successful steps indicate that the model is (at least locally) an adequate approximation of the objective such that the penalty parameter is decreased in order to allow for longer steps. For unsuccessful iterations we proceed exactly the opposite way. Readers familiar with trust region methods might see that one can interpret the penalty parameter $\sigma_k$ as inversely proportional to the trust region radius $\delta_k$.


 \begin{algorithm*}[tb]
   \caption{Sub-sampled Cubic Regularization (SCR)}
   \label{alg:src}
\begin{algorithmic}[1]
   \STATE {\bfseries Input:} \\ 
   $\quad$ Starting point $\x_0 \in \R^d$ (e.g~$\x_0 = {\bf 0}$) \\
   $\quad \gamma >1, 1>\eta_2>\eta_1>0$, and $\sigma_0>0$
   \FOR{$k=0,1,\dots,\text{until convergence}$}
   \STATE Sample gradient $\g_k$ and Hessian $\Hm_k$ according to Eq.~\ref{eq:gradient_sampling_scheme} \& Eq.~\ref{eq:hessian_sampling_scheme} respectively
   \STATE Obtain $\s_k$ by solving $m_k(\s_k)$ (Eq. ~\ref{eq:subproblem}) such that A\ref{a:approx_model_min} holds
   \STATE Compute $f(\x_k + \s_k)$ and 
	\begin{equation}
	\rho_k=\dfrac{f(\x_k)-f(\x_k + \s_k)}{f(\x_k)-m_k(\s_k)}
	\end{equation}
	\STATE Set
	\begin{equation}
	\x_{k+1} = \begin{cases}
	\x_k + \s_k & \text{ if } \rho_k \geq \eta_1\\
	\x_k & \text{ otherwise}
	\end{cases}
	\end{equation}
	\STATE Set
	\begin{equation} \label{eq:sigma_update}
	\sigma_{k+1}= \begin{cases}
	\max\{\min\{\sigma_k,\norm{\g_k}\},\epsilon_m \} & \text{ if } \rho_k>\eta_2 \text{ (very successful iteration)}\\
	\sigma_k & \text{ if } \eta_2\geq\rho_k\geq \eta_1 \text{ (successful iteration)}\\
	\gamma\sigma_k & \text{ otherwise}\text{ (unsuccessful iteration)},
	\end{cases}
	\end{equation}
	where $\epsilon_m \approx10^{-16}$ is the relative machine precision.
   \ENDFOR
\end{algorithmic}
\end{algorithm*}

\subsection{Exact model minimization}
\label{sec:exact_minimization}

Solving Eq.~\ref{eq:subproblem} requires minimizing an unconstrained non-convex problem that may have isolated local minima. As shown in~\cite{cartis2011adaptive} the global model minimizer $\s_k^*$ is characterized by following systems of equations,
\begin{equation}\label{eq:global_minimizer}
(\Bm_k + \lambda_k^* \Im) \s_k^*=-\g_k, \: \lambda_k^* = \sigma_k \norm{\s_k^*} , (\Bm_k + \lambda_k^* \Im) \succeq 0.
\end{equation}

In order to find a solution we can express $\s_k^* := \s_k(\lambda_k^*) = -(\Bm_k + \lambda^*_k \Im)^{-1} \g_k$, apply this in the second equation of (\ref{eq:global_minimizer}) and obtain a univariate, nonlinear equation in $\lambda_k$
\begin{equation} \label{eq:root_problem}
\norm{-(\Bm_k + \lambda^*_k \Im)^{-1} \g_k}-\frac{\lambda_k^*}{\sigma_k}=0.
\end{equation}
Furthermore, we need $\lambda^*_k \geq \max \{-\lambda_1(\Bm_k),0\}$, where $\lambda_1(\Bm_k)$ is the leftmost eigenvalue of $\Bm_k$, in order to guarantee the semi-positive definiteness of $(\Bm_k + \lambda^*_k \Im)$.

Thus, computing the global solution of $m_k$ boils down to finding the root of Eq.~\ref{eq:root_problem} in the above specified range of $\lambda_k$. The problem can be solved by Newton's method, which involves factorizing $\Bm_k + \lambda_k \Im$ for various $\lambda_k$ and is thus prohibitively expensive for large problem dimensions $d$. See Section 6.2 in~\cite{cartis2011adaptive} for more details.
In the following Section we instead explore an approach to \textit{approximately} minimize the model while retaining the convergence guarantees of the exact minimization.

\subsection{Approximate model minimization}
\label{sec:approximate_minimization}

\cite{cartis2011adaptive} showed that it is possible to retain the remarkable properties of the cubic regularization algorithm with an inexact model minimizer. A necessary condition is that $\s_k$ satisfies the two requirements stated in A\ref{a:approx_model_min}.

\begin{assumption}[Approximate model minimizer] \label{a:approx_model_min}
\begin{eqnarray}
\s_k^\intercal \g_k + \s_k^\intercal \Bm_k \s_k + \sigma_k \norm{\s_k}^3=0 \label{eq:3.11}\\
\s_k^\intercal \Bm_k \s_k + \sigma_k \norm{\s_k}^3 \geq 0 \label{eq:3.12}
\end{eqnarray}
\end{assumption}
Note that the first equation is equal to $\nabla_s m_k(\s_k)^\intercal \s_k=0$ and the second to $\s_k^\intercal  \nabla^2_s m_k(\s_k) \s_k\geq0$.

As shown in~\citep{cartis2011adaptive} Lemma 3.2, the global minimizer of $m_k(\s_k)$ in a Krylov subspace $\mathcal{K}_k:=\text{span}\{\g_k,\Hm_k\g_k,\Hm_k^2\g_k,...\}$ satisfies this assumption independent of the subspace dimension. This comes in handy, as minimizing $m_k$ in the Krylov subspace only involves factorizing a tri-diagonal matrix, which can be done at the cost of $O(d)$. However, a Lanczos-type method must be used in order to build up an orthogonal basis of this subspace which typically involves one matrix-vector product ($O((2d-1)n)$) per additional subspace dimension (see Chapter 5 in \citep{conn2000trust} for more details).

Thus, in order to keep the per iteration cost of \methodname low and in accordance to ARC, we apply the following termination criterion to the Lanczos process in the hope to find a suitable trial step before $\mathcal{K}_k$ is of dimensionality $d$.

\begin{assumption}[Termination Criteria] \label{a:TC}
For each outer iteration $k$, assume that the Lanczos process stops as soon as some Lanczos iteration $i$ satisfies the criterion 
\begin{equation}\label{eq:TC}
\text{TC: }\norm{\nabla m_k(\s_{i,k})} \leq \theta_k \norm{\g_k},
\end{equation}
where $\theta_k=\kappa_\theta \min(1,\norm{\s_{i,k}}),\: \kappa_\theta \in (0,1)$.
\end{assumption} 

However, we argue that especially for high dimensional problems, the cost of the Lanczos process may significantly slow down cubically regularized methods and since this cost increases linearly in $n$, carefully sub-sampled versions are an attractive alternative.

\section{Theoretical analysis}
\label{sec:analysis}

In this section, we provide the convergence analysis of SCR. For the sake of brevity, we assume Lipschitz continuous Hessians right away but note that a superlinear local convergence result as well as the global first-order convergence theorem can both be obtained without the former assumption.

First, we lay out some critical assumptions regarding the gradient and Hessian approximations. Second, we show that one can theoretically satisfy these assumptions with high probability by sub-sampling first- and second-order information. Third, we give a condensed convergence analysis of \methodname which is widely based on \citep{cartis2011adaptive}, but adapted for the case of stochastic gradients. There, we show that the local and global convergence properties of ARC can be retained by sub-sampled versions at the price of slightly worse constants.
\subsection{Assumptions}
\begin{assumption}[Continuity]\label{a:continuity}
The functions $f_i \in C^2(\mathbb{R}^d)$, $g_i$ and $H_i$ are Lipschitz continuous for all $i$, with Lipschitz constants $\kappa_f, \kappa_g$ and $\kappa_H$ respectively.
\end{assumption}
By use of the triangle inequality, it follows that these assumptions hold for all $\g$ and $\Hm$, independent of the sample size. Furthermore, note that the Hessian and gradient norms are uniformly bounded as a consequence of A\ref{a:continuity}.

In each iteration, the Hessian approximation $\Bm_k$ shall satisfy condition AM.4 from \citep{cartis2011adaptive}, which we restate here for the sake of completeness.
\begin{assumption}[Sufficient Agreement of $\Hm$ and $\Bm$]\label{a:Strong_agreement_H}
\begin{equation}\label{eq:Strong_agreement_H}
\norm{(\Bm_k-\Hm(\x_k)) \s_k} \leq C \norm{\s_k}^2, \:\forall k\geq 0, C>0.
\end{equation}
\end{assumption}
We explicitly stress the fact that this condition is stronger than the well-known Dennis Mor\'e Condition. 
While quasi-Newton approximations satisfy the latter, there is no theoretical guarantee that they also satisfy the former \cite{cartis2011adaptive}. Furthermore, any sub-sampled gradient shall satisfy the following condition.
\begin{assumption}[Sufficient Agreement of $\nabla f$ and $g$]\label{a:Strong_agreement_g}
\begin{equation}\label{eq:Strong_agreement_g}
\norm{\nabla f(\x_k) - \g(\x_k)} \leq M \norm{\s_k}^2, \:\forall k\geq 0,\: M>0.
\end{equation}
\end{assumption}

\subsection{Sampling Conditions}
Based on probabilistic deviation bounds for random vectors and matrices\footnote{These bounds have lately become popular under the name of \textit{concentration inequalities}. Unlike classic limit theorems, such as the Central Limit Theorem, concentration inequalities are specifically attractive for application in machine learning because of their non-asymptotic nature.}, we now present sampling conditions that guarantee sufficient steepness and curvature information in each iteration $k$.
In particular, the Bernstein inequality gives exponentially decaying bounds on the probability of a random variable to differ by more than $\epsilon$ from its mean for any fixed number of samples. We use this inequality to upper bound the $\ell_2$-norm distance $\|\nabla f-\g\|$, as well as the spectral-norm distance $\|\Bm-\Hm\|$ by quantities involving the sample size $|S|$. By applying the resulting bounds in the sufficient agreement assumptions (A\ref{a:Strong_agreement_H} \& A\ref{a:Strong_agreement_g}) and re-arranging for $|S|$, we are able to translate the latter into concrete sampling conditions.

\subsubsection{Gradient Sampling}
As detailed in the Appendix, the following Lemma arises from the Vector Bernstein Inequality.
\begin{lemma}[Gradient deviation bound]\label{l:gradient_deviation}
Let the sub-sampled gradient $\g_k$ be defined as in Eq.~\ref{eq:subproblem}. For $\epsilon\leq2\kappa_f$ we have with probability $(1-\delta)$ that
\begin{equation}\label{eq:g_deviation}
\norm{\g(\x_k)-\nabla f(\x_k)} \leq 4\sqrt{2} \kappa_f \sqrt{\frac{\log ((2d)/\delta)+1/4}{|S_{g,k}|}}.
\end{equation}
\end{lemma} 
It constitutes a non-asymptotic bound on the deviation of the gradient norms that holds with high probability. Note how the accuracy of the gradients increases in the sample size. This bound yields the following condition. 
\begin{framed}
\begin{theorem}[Gradient Sampling]\label{t:grad_sampling}
If 
\begin{equation}\label{eq:gradient_sampling_scheme}
|S_{g,k}|\geq \dfrac{32\kappa_f^2\left(\log\left((2d)/\delta\right)+1/4\right)}{M^2\norm{\s_k}^4},\: M\geq 0, \forall k\geq 0
\end{equation}
then $\g_k$ satisfies the sufficient agreement condition A\ref{a:Strong_agreement_g} with probability $(1-\delta)$.
\end{theorem}
\end{framed}

\subsubsection{Hessian Sampling}
In analogy to the gradient case, we use the matrix version of Bernstein's Inequality to derive the following Lemma.
\begin{lemma}[Hessian deviation bound]\label{l:Hessian_deviation}
Let the sub-sampled Hessian $\Bm$ be defined as in Eq.~\ref{eq:subproblem}. For $\epsilon \leq 4\kappa_g$ we have with probability $(1-\delta)$ that
\begin{equation}\label{eq:hessian_sampling_scheme}
\norm{\Bm(\x_k)-\Hm(\x_k)} \leq 4\kappa_g \sqrt{\frac{\log(2d/\delta)}{|S_{B,k}|}},
\end{equation}
\end{lemma}

This, in turn, can be used to derive a Hessian sampling condition that is guaranteed to satisfy the sufficient agreement condition (A\ref{a:Strong_agreement_H}) with high probability.
\begin{framed}
\begin{theorem}[Hessian Sampling]\label{t:hessian_sampling}
If
\begin{equation}\label{eq:hessian_sampling_scheme}
|S_{B,k}| \geq \frac{16\kappa_g^2\log(2d/\delta)}{(C\norm{\s_k})^2},\: C\geq 0 \text{, and } \forall k\geq 0
\end{equation}
then $\Bm_k$ satisfies the strong agreement condition A\ref{a:Strong_agreement_H} with  probability $(1-\delta)$.
\end{theorem}
\end{framed}

As expected, the required sample size grows in the problem dimensionality $d$ and in the Lipschitz constants $\kappa_f$ and $\kappa_g$. Finally, as outlined in the Appendix (Lemma \ref{l:eventu_full_samples}), the samples size is eventually equal to the full sample size $n$ as \methodname converges and thus we have \begin{equation} \label{eq:BtoH}
\g \rightarrow \nabla f \text{ as well as } \Bm \rightarrow \Hm \text{ as } k \rightarrow \infty.
\end{equation}

\subsection{Convergence Analysis}
The entire analysis of cubically regularized methods is prohibitively lengthy and we shall thus establish only the crucial properties that ensure global, as well as fast local convergence and improve the worst-case complexity of these methods over standard trust region approaches. Next to the cubic regularization term itself, these properties arise mainly from the penalty parameter updates and step acceptance criteria of the ARC framework, which give rise to a good relation between regularization and stepsize. Further details can be found in~\cite{cartis2011adaptive}.

\subsubsection{Preliminary Results}
First, we note that the penalty parameter sequence $\{\sigma_k\}$ is guaranteed to stay within some bounded positive range, which is essentially due to the fact that \methodname is guaranteed to find a successful step as soon as the penalty parameter exceeds some critical value $\sigma_{sup}$.
\vspace{5mm}
\begin{lemma}[Boundedness of $\sigma_k$]\label{l:bounds_on_sigma}
Let A\ref{a:continuity}, A\ref{a:Strong_agreement_H} and A\ref{a:Strong_agreement_g} hold. Then
\begin{equation}
\sigma_k \in [\sigma_{\inf}, \sigma_{\sup} ],\: \forall k \geq 0,
\end{equation}
where $\sigma_{\inf}$ is defined in Step 7 of Algorithm \ref{alg:src} and 
\begin{equation}\label{eq:upper_bound_on_sigma}
\sigma_{\text{sup}} := \{\sigma_0,\frac{3}{2}\gamma_2 (2M+C+\kappa_g)\}.
\end{equation}
\end{lemma}

Furthermore, for any successful iteration the objective decrease can be directly linked to the model decrease via the step acceptance criterion in Eq.~\ref{eq:sigma_update}. The latter, in turn, can be shown to be lower bounded by the stepsize which combined gives the following result.
\vspace{5mm}
\begin{lemma}[Sufficient function decrease] \label{l:lower_bound_function_decrease}
Suppose that $\s_k$ satisfies A\ref{a:approx_model_min}. Then, for all successful iterations $k\geq0$
\begin{equation}\label{eq:lower_bound_model_decrease}
\begin{aligned}
f(\x_k)-f(\x_{k+1})&\geq \eta_1(f(\x_k)-m(\s_k))\\
&\geq \frac{1}{6}\eta_1\sigma_{\inf} \norm{\s_k}^3
\end{aligned}
\end{equation}
\end{lemma}
Finally, the termination criterion (\ref{eq:TC}) also guarantees step sizes that do not become too small compared to the respective gradient norm which leads to the following Lemma.

\begin{lemma}[Sufficiently long steps]\label{l:lower_bound_on_s}
Let  A\ref{a:continuity}, A\ref{a:Strong_agreement_H} and A\ref{a:Strong_agreement_g} hold. Furthermore, assume the termination criterion TC (A\ref{a:TC}) and suppose that $\x_k\rightarrow \x^*, \text{as } k\rightarrow \infty$. 
Then, for all sufficiently large successful iterations, $\s_k$ satisfies
\begin{equation}\label{eq:better_lower_bound_on_s}
\: \norm{\s_k} \geq \kappa_s\: \sqrt{\norm{\nabla f(\x_{k+1})}}
\end{equation}
where $\kappa_s$ is the positive constant
\begin{equation}\label{eq:kappa_s}
\kappa_s = \sqrt{ \frac{1-\kappa_{\theta}}{\frac{1}{2}\kappa_g+(1+\kappa_{\theta}\kappa_g)M+C+\sigma_{\sup}+\kappa_{\theta}\kappa_g}}.
\end{equation}   
\end{lemma}


\subsubsection{Local convergence}
We here provide a proof of local convergence for any sampling scheme that satisfies the conditions presented in Theorem \ref{t:grad_sampling} and Theorem \ref{t:hessian_sampling} as well as the additional condition that the sample size does not decrease in unsuccessful iterations. We show that such sampling schemes eventually yield exact gradient and Hessian information. Based upon this observation, we obtain the following local convergence result (as derived in the Appendix).\begin{framed}
\begin{theorem}[Quadratic local convergence]
\label{t:quadratic_convergence_exp}
Let A\ref{a:continuity} hold and assume that $\g_k$ and $\Bm_k$ are sampled such that \ref{eq:gradient_sampling_scheme} and \ref{eq:hessian_sampling_scheme} hold and $|S_{g,k}|$ and $|S_{B,k}|$ are not decreased in unsuccessful iterations. Furthermore, let $s_k$ satisfy A\ref{a:approx_model_min} and
\begin{equation}\label{temp:limit_exp}
\x_k\rightarrow \x^*,\: \text{as } k \rightarrow \infty,
\end{equation}
where $\Hm(\x^*)$ is positive definite.  Moreover, assume the stopping criterion TC (A\ref{a:TC}). Then,
\begin{equation}\label{eq:quadratic_x_exp}
\dfrac{\norm{\x_{k+1}-\x^*}}{\norm{\x_{k}-\x^*}^2} \leq c,\: c>0 \text{ as } k\rightarrow \infty \;(w.h.p.).
\end{equation}
That is, $\x_k$ converges in q-quadratically to $\x^*$ as $k\rightarrow \infty$ with high probability.
\end{theorem}
\end{framed}
\subsubsection{Global convergence to first-order critical point}\label{t:foc}
Lemma \ref{l:bounds_on_sigma} and \ref{l:lower_bound_function_decrease} allow us to lower bound the function decrease of a successful step in terms of the  \textit{full} gradient $\nabla f_k$ (as we will shorty detail in Eq.~\ref{eq:sufficient_function_decrease}). Combined with Lemma \ref{l:bounds_on_sigma}, this allows us to give deterministic global convergence guarantees using only stochastic first order information.
\begin{framed}
\begin{theorem}[Convergence to 1st-order Critical Points]\label{t:1st_order_guarantee}
Let A\ref{a:approx_model_min}, A\ref{a:continuity}, A\ref{a:Strong_agreement_H} and A\ref{a:Strong_agreement_g} hold. Furthermore, let $\{f(\x_k)\}$ be bounded below by some $f_{\inf} >-\infty$. Then
\begin{equation}
\lim_{k\rightarrow \infty} \norm{\nabla f(\x_k)} = 0
\end{equation}
\end{theorem}
\end{framed}
\subsubsection{Global convergence to second-order critical point}
Unsurprisingly, the second-order convergence guarantee relies mainly on the use of second-order information so that the stochastic gradients do neither alter the result nor the proof as it can be found in Section 5 of \cite{cartis2011adaptive}. We shall restate it here for the sake of completeness.
\begin{framed}

\begin{theorem}[Second-order global convergence]\label{t:2nd_order_guarantee}
Let  A\ref{a:continuity}, A\ref{a:Strong_agreement_H} and A\ref{a:Strong_agreement_g} hold. Furthermore, let $\{ f(\x_k)\}$ be bounded below by $f_{\inf}$ and $\s_k$ be a global minimizer of $m_k$ over a subspace $\mathcal{L}_k$ that is spanned by the columns of the $d\times l$ orthogonal matrix $\Qm_k$. As $\Bm \rightarrow \Hm$ asymptotically (Eq. \ref{eq:BtoH}), any subsequence of negative leftmost eigenvalues $\{\lambda_{\min}(\Qm_k^\intercal \Hm(\x_k)\Qm_k)\}$ converges to zero for sufficiently large, successful iterations. Hence
\begin{equation}\label{eq:lambda_H_limit}
\lim_{k\in \mathcal{S}} \inf_{k\rightarrow \infty} \lambda_{\min}(\Qm_k^\intercal \Hm(\x_k)\Qm_k) \geq 0.
\end{equation}
Finally, if $\Qm_k$ becomes a full orthogonal basis of $\mathbb{R}^d$ as $k\rightarrow \infty$, then any limit point of the sequence of successful iterates $\{\x_k\}$ is second-order critical (provided such a limit point exists).
\end{theorem}
\end{framed}

\subsubsection{Worst-case iteration complexity} \label{s:wcc}
For the worst-case analysis we shall establish the two disjoint index sets $\mathcal{U}_j$ and $\mathcal{S}_j$, which represent the un- and successful SCR iterations that have occurred up to some iteration $j>0$, respectively.
As stated in Lemma \ref{l:bounds_on_sigma} the penalty parameter $\sigma_k$ is bounded above and hence \methodname  may only take a limited number of consecutive unsuccessful steps. As a consequence, the total number of unsuccessful iterations is at most a problem dependent constant times the number of successful iterations. 
\begin{lemma}[Number of unsuccessful iterations]\label{l:number_of_unsuccessful_steps}
For any fixed $j\geq 0$, let Lemma \ref{l:bounds_on_sigma} hold. Then we have that
\begin{equation}\label{eq:number_of_unsuccessful_steps}
|\mathcal{U}_j| \leq \left\lceil (|\mathcal{S}_j|+1)\dfrac{\log(\sigma_{\sup})-\log(\sigma_{\inf})}{\log(\eta_1)} \right\rceil.
\end{equation}
\end{lemma}

Regarding the number of successful iterations we have already established the two key ingredients: (i) a sufficient function decrease in each successful iteration (Lemma \ref{l:lower_bound_function_decrease}) and (ii) a step size that does not become too small compared to the respective gradient norm (Lemma \ref{l:lower_bound_on_s}), which is essential to driving the latter below $\epsilon$ at a fast rate. Combined they give rise to the  guaranteed function decrease for successful iterations
\begin{equation}
f(\x_k)-f(\x_{k+1})\geq \frac{1}{6}\eta_1\sigma_{\inf}\kappa_s^3\:\norm{\nabla f(\x_{k+1})}^{3/2},
\label{eq:sufficient_function_decrease}
\end{equation}
which already contains the power of 3/2 that appears in the complexity bound. Finally, by summing over all successful iterations one obtains the following, so far best know, worst-case iteration bound to reach $\epsilon$ first-order criticality.

\begin{framed}
\begin{theorem}[First-order worst-case complexity]\label{t:fowcc}
Let A\ref{a:approx_model_min}, A\ref{a:continuity}, A\ref{a:Strong_agreement_H} and A\ref{a:Strong_agreement_g} hold. Furthermore, be $\{f(\x_k)\}$ bounded below by $f_{\inf}$ and TC applied (A\ref{a:TC}). Then, for $\epsilon>0$ the total number of iterations \methodname  takes to generate the first iterate $j$ with $\norm{\nabla f(\x_{j+1})}\leq \epsilon$, and assuming $\epsilon \leq 1$, is
\begin{equation}\label{eq:number_total_iterations}
j\leq \left\lceil (1+\kappa_i)(2+\kappa_j) \epsilon^{-3/2} \right\rceil, 
\end{equation}
where 
\begin{equation}
\kappa_i=6\dfrac{f(\x_0)-f_{\inf}}{\eta_1\sigma_{\inf}\kappa_s^3}  \text{ and }\kappa_j = \dfrac{\log(\sigma_{\sup})-\log(\sigma_{\inf})}{\log(\eta_1)}
\end{equation}
\end{theorem}
\end{framed}
Note that the constants $\kappa_i$ and $\kappa_j$ involved in this upper bound both increase in the gradient inaccuracy $M$ and the Hessian inaccuracy $C$ (via $\kappa_s$ and $\sigma_{\sup}$), such that more inaccuracy in the sub-sampled quantities may well lead to an increased overall number of iterations.

Finally, we want to point out that similar results can be established regarding a second-order worst-case complexity bound similar to Corollary 5.5 in \citep{cartis2011adaptive2}, which we do not prove here for the sake of brevity.


\section{Experimental results}
In this section we present experimental results on real-world datasets where $n \gg d \gg 1$. They largely confirm the analysis derived in the previous section. Please refer to the Appendix for more detailed results and experiments on higher dimensional problems.

\begin{figure*}[t!]\label{fig:exp_results}
	\begin{center}
          \begin{tabular}{@{}c@{\hspace{2mm}}c@{\hspace{2mm}}c@{\hspace{2mm}}c@{}}
          \vspace{-5.5pt}
            \includegraphics[width=0.29\linewidth]{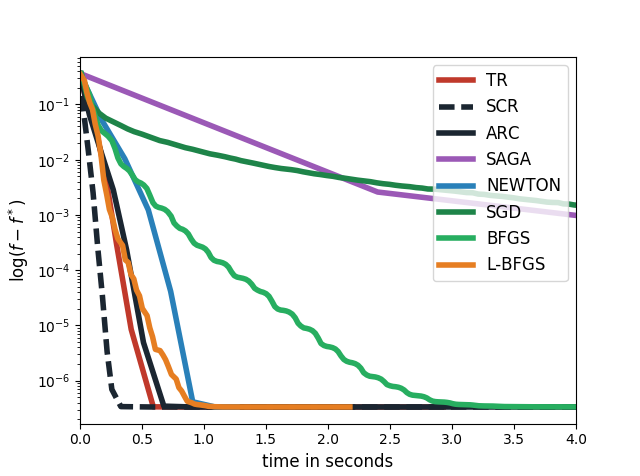} &
            \includegraphics[width=0.29\linewidth]{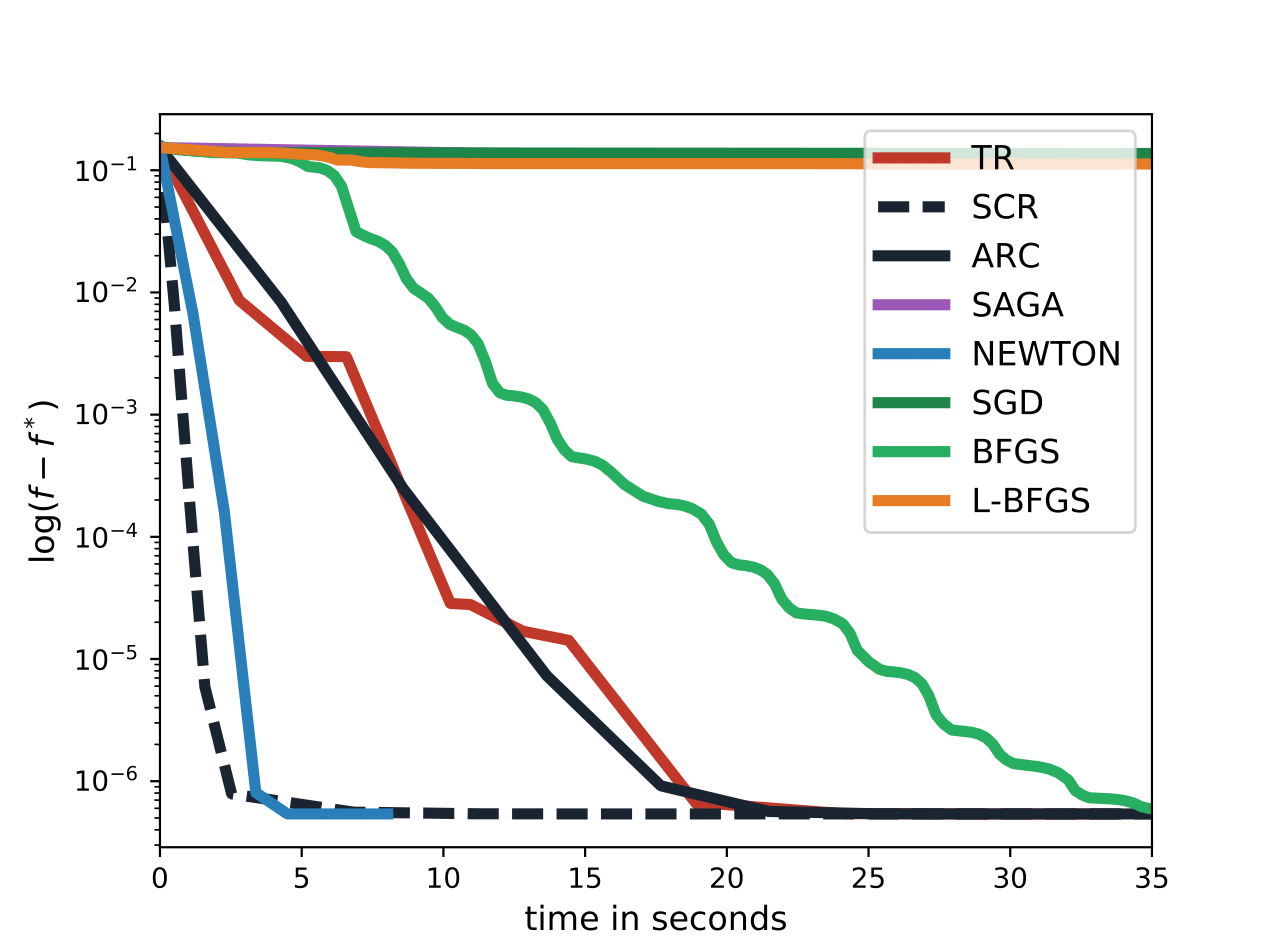} & 
             \includegraphics[width=0.29\linewidth]{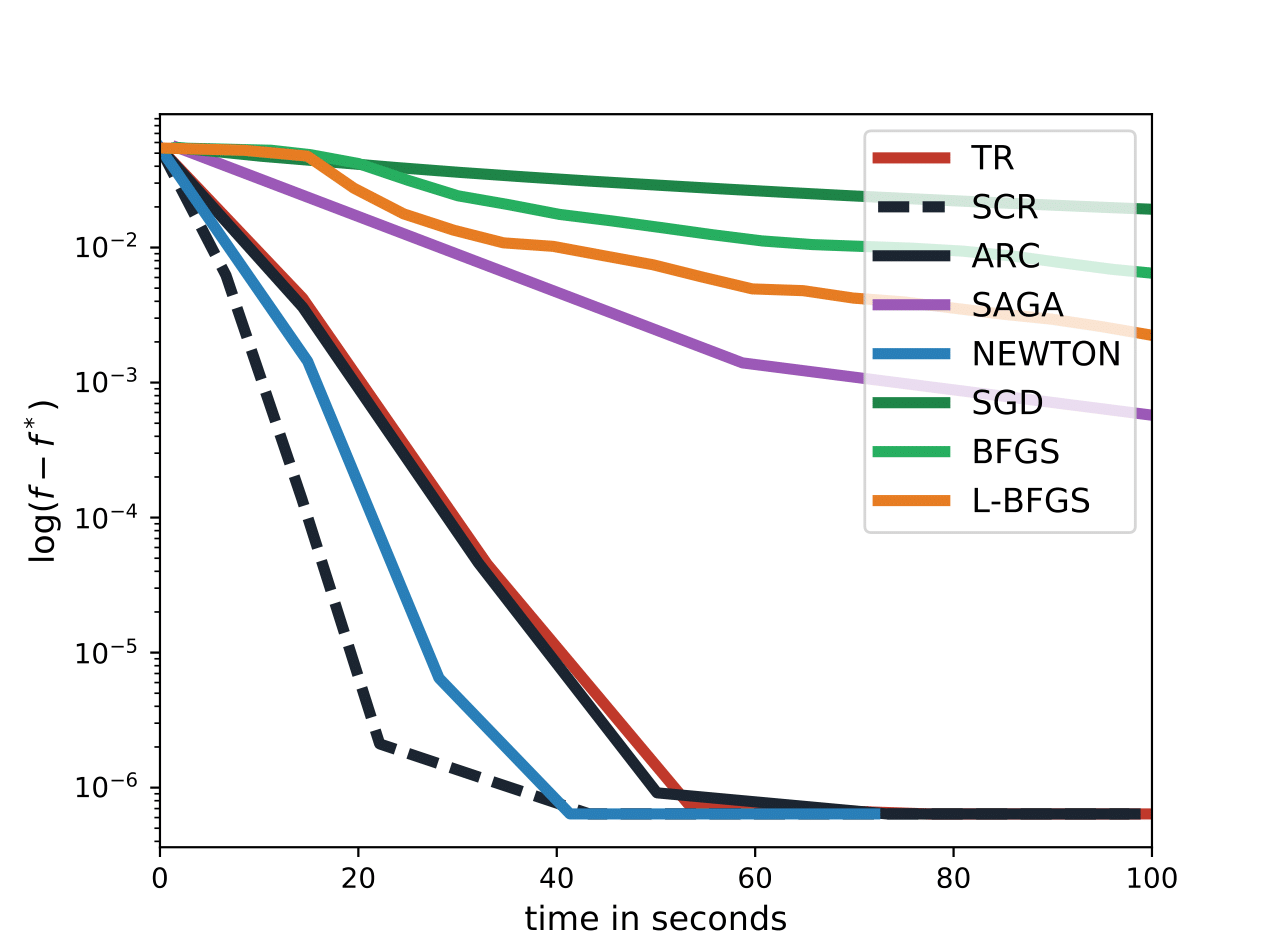}  \\ 
             \includegraphics[width=0.29\linewidth]{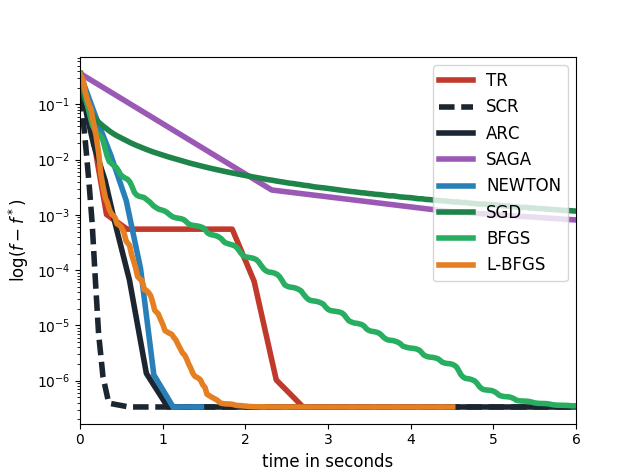} &
            \includegraphics[width=0.29\linewidth]{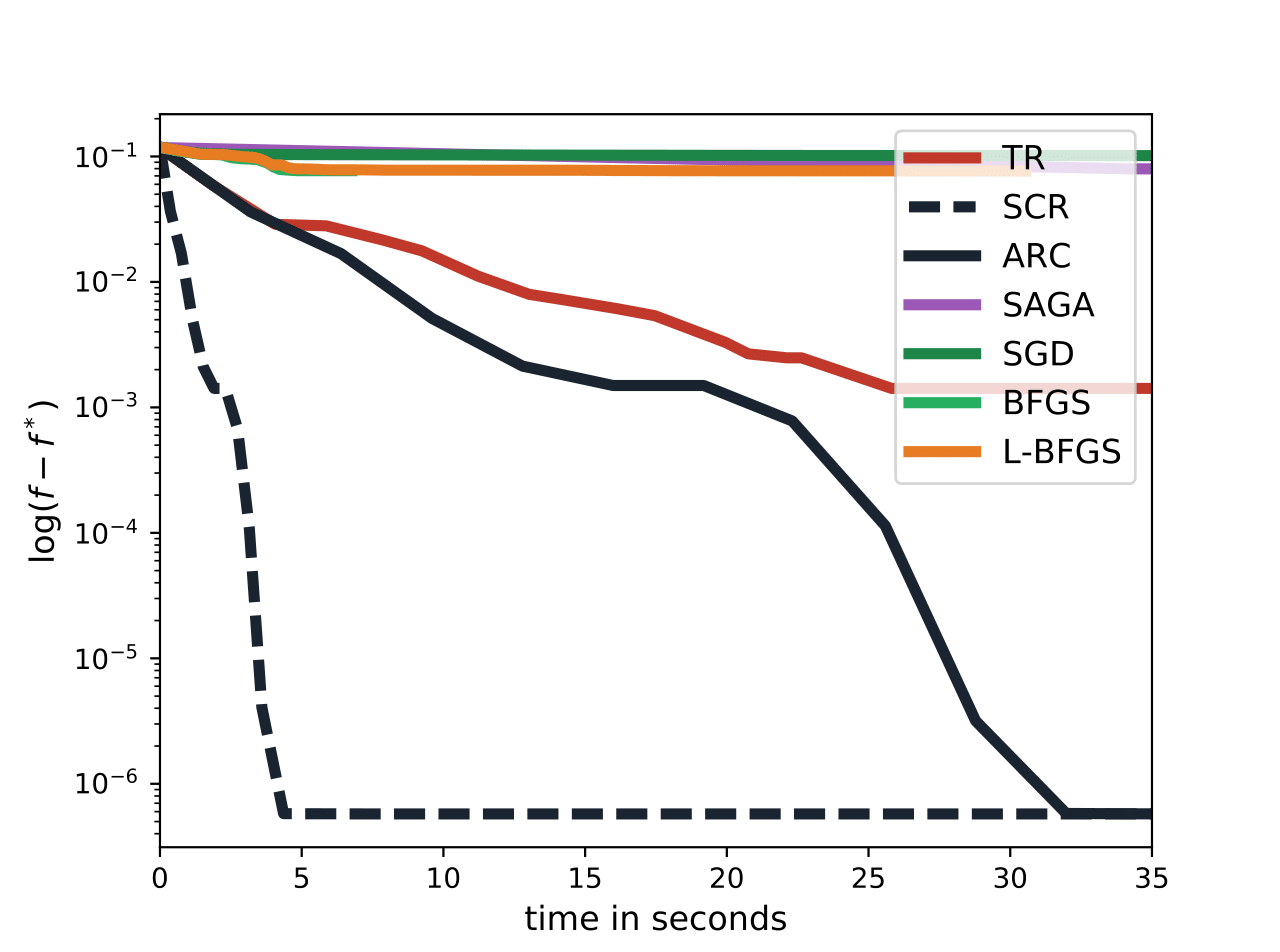} & 
             \includegraphics[width=0.29\linewidth]{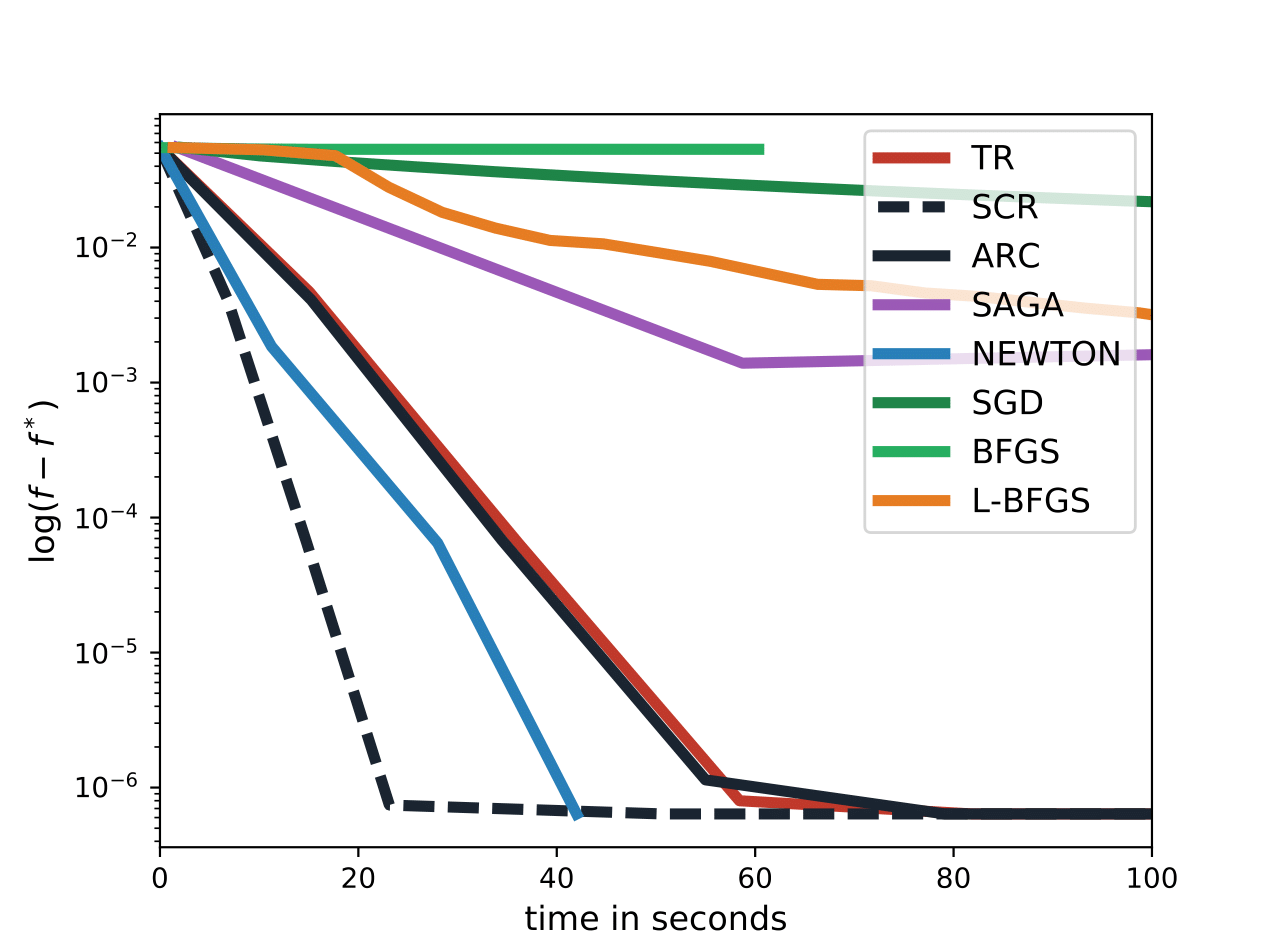}  \\  \vspace{-2pt}
         	1. \footnotesize{{ A9A (n=32561, d=123)}} &
            2. \footnotesize{{ COVTYPE (n=581012, d=54)}} & 
            3. \footnotesize{{ HIGGS (n=11000000, d=28)}} \vspace{-2pt}
	  \end{tabular}
          \caption{Top (bottom) row shows log suboptimality of convex (non-convex) regularized logistic regressions over time (avg. of 10 runs).}
          \label{fig:results}

	\end{center}
\end{figure*}

\subsection{Practical implementation of \methodname}
We implement \methodname as stated in Algorithm \ref{alg:src} and note the following details. Following~\cite{erdogdu2015convergence}, we require the sampling conditions derived in Section~\ref{sec:analysis} to hold with probability $O(1-1/d)$, which yields the following practically applicable sampling schemes
\begin{equation}\label{eq:sampling_scheme}
\begin{aligned}
&|S_{k,H}| \geq \frac{36\kappa_g^2\log(d)}{(C\norm{\s_k})^2},\: C>0, \: \forall k>0\\
&|S_{k,g}| \geq \frac{32\kappa_f^2(\log(d)+1/4)}{M^2\|\s_k\|^4},\: M>0, \: \forall k>0.
\end{aligned}
\end{equation}
The positive constants $C$ and $M$ can be used to scale the sample size to a reasonable portion of the entire dataset and can furthermore be used to offset $\kappa_g$ and $\kappa_f$, which are generally expensive to obtain. 
 
However, when choosing $|S|$ for the current iteration $k$, the stepsize $\s_k$ is yet to be determined.
Based on the Lipschitz continuity of the involved functions, we argue that the previous stepsize is a fair estimator of the current one and this is confirmed by experimental results. Finally, we would like to point out that the sampling schemes derived in Eq.~\ref{eq:sampling_scheme} gives our method a clear edge over sampling schemes that do not take any iteration information into account, e.g. linearly or geometrically increased samples.

\subsection{Baselines and datasets}
We compare \methodname{}to various optimization methods presented in Section~\ref{sec:related_work}. This includes SGD (with constant step-size), SAGA, Newton's method, BFGS, L-BFGS and ARC. More details concerning the choice of the hyper-parameters are provided in the appendix. We ran experiments on the  datasets \textit{a9a}, \textit{covtype} and \textit{higgs} (see details in the appendix). We experimented with a binary logistic regression model with two different regularizers: a standard $\ell_2$ penalty $\lambda \| x \|^2$, and a non-convex regularizer $\lambda \sum_{i=1}^d x_{(i)}^2/ \left(1+ x_{(i)}^2 \right)$ (see~\cite{reddi2016fast}).
\subsection{Results}
The results in Figure \ref{fig:results} confirm our intuition that \methodname can reduce ARCs computation time without losing its global convergence property. Newton's method is the closest in terms of performance. However, it suffer heavily from an increase in $d$ as can be seen by additional results provided in the appendix. Furthermore, it cannot optimize the non-convex version of \textit{covtype} due to a singular Hessian. Notably, BFGS terminates early on the non-convex \textit{higgs} dataset due to a local saddle point. Finally, the high condition number of \textit{covtype} has a significant effect on the performance of SGD, SAGA and L-BFGS.


\section{Conclusion}

In this paper we proposed a sub-sampling technique to estimate the gradient and Hessian in order to construct a cubic model analogue to trust region methods. We show that this method exhibits the same convergence properties as its deterministic counterpart, which are the best known worst-case convergence properties on non-convex functions. Our proposed method is especially interesting in the large scale regime when $n \gg d$. Numerical experiments on both real and synthetic datasets demonstrate the performance of the proposed algorithm which we compared to its deterministic variant as well as more classical optimization methods. As future work we would like to explore the adequacy of our method to train neural networks which are known to be hard to optimize due to the presence of saddle points.

\bibliography{scr}
\bibliographystyle{icml2017}


\newpage
\appendix
\onecolumn

\section{Appendix}
\label{App:details}

\subsection{Concentration Inequalities and Sampling Schemes}
For the sake of simplicity we shall drop the iteration subscript $k$ in the following results of this section.
\subsubsection{Gradient Sampling}
First, we extend the Vector Bernstein inequality as it can be found in \cite{gross2011recovering} to the \textit{average} of independent, zero-mean vector-valued random variables.
\begin{framed}
\begin{lemma}[Vector Bernstein Inequality]
Let $\x_1,\ldots,\x_n$ be independent vector-valued random variables with common dimension $d$ and assume that each one is centered, uniformly bounded and also the variance is bounded above:
\begin{equation*}
\e{\x_i}=0 \text{ and } \norm{\x_i}_2 \leq \mu \text{ as well as } \e{\norm{\x_i}^2}\leq \sigma^2
\end{equation*}
Let
\begin{equation*}
\z=\frac{1}{n}\sum_{i=1}^n \x_i.
\end{equation*}
Then we have for $0<\epsilon <\sigma^2/\mu$
\begin{equation} \label{eq:final_bernstein_vector}
P\left(\norm{\z}\geq \epsilon\right) \leq \exp\left(-n\cdot \frac{\epsilon^2}{8\sigma^2}+\frac{1}{4}\right)
\end{equation}
\end{lemma}
\end{framed}
\textit{Proof:}
Theorem 6 in \cite{gross2011recovering} gives the following Vector Bernstein inequality for independent, zero-mean vector-valued random variables
\begin{equation}\label{eq:original_vector_bernstein}
P\left(\norm{\sum_{n=1}^n \x_i}\geq t + \sqrt{V}\right) \leq \exp\left(- \frac{t^2}{4V}\right),
\end{equation}
where $V=\sum_{i=1}^n \e{\norm{\x_i}^2}$ is the sum of the variances of the centered vectors $\x_i$.

First, we shall define $\epsilon = t+\sqrt{V}$, which allows us to rewrite the above equation as

\begin{equation}
P\left(\norm{\sum_{i=1}^n \x_i}\geq \epsilon\right) \leq \exp\left(-\frac{1}{4} \left(\frac{\epsilon-\sqrt{V}}{\sqrt{V}}\right)^2\right)=\exp\left(-\frac{1}{4} \left(\frac{\epsilon}{\sqrt{V}}-1\right)^2 \right).
\end{equation}
Based on the observation that
\begin{equation}
\begin{aligned}
&-\frac{1}{4} \left(\frac{\epsilon}{\sqrt{V}}-1\right)^2 \leq - \frac{1}{4}\left(\frac{\epsilon^2}{2V}\right)+\frac{1}{4} \\
\Leftrightarrow \: &-\frac{\epsilon^2}{V}+2\frac{\epsilon}{\sqrt{V}}-1\leq -\frac{\epsilon^2}{2V}+1\\
\Leftrightarrow \: & 0 \leq \frac{\epsilon^2}{2V}-2\frac{\epsilon}{\sqrt{V}}+2 \\
\Leftrightarrow \: & 0 \leq \left(\frac{\epsilon}{\sqrt{2V}}-\sqrt{2}\right)^2
\end{aligned}
\end{equation}
always holds, we can formulate a slightly weaker Vector Bernstein version as follows

\begin{equation}\label{eq:modified_vector_bernstein}
P\left(\norm{\sum_{i=1}^n \x_i}\geq \epsilon\right) \leq \exp\left(-\frac{\epsilon^2}{8V}+\frac{1}{4}\right).
\end{equation}

Since the individual variance is assumed to be bounded above, we can write
\begin{equation}
V=\sum_{i=1}^n \e{\norm{\x_i}^2} \leq n \sigma^2.
\end{equation}
This term also constitutes an upper bound on the variance of $\y=\sum_{i=1}^n \x_i$, because the $\x_i$ are independent and thus uncorrelated 
. However, $\z=\frac{1}{n}\sum_{i=1}^n \x_i$ and we must account for the averaging term. Since the $\x_i$ are centered we have $\e{\z}=0$, and thus
\begin{equation}\label{temp:bernstein_grad1}
\begin{aligned}
Var(\z)=\e{\norm{\z-\e{\z}}^2}=\e{\norm{\z}^2} =\e{\norm{\frac{1}{n}\sum_{i=1}^n \x_i}^2}=\frac{1}{n^2}\e{\left(\sum_{i=1}^n \x_i\right)^\intercal\left(\sum_{j=1}^n \x_j\right)}\\
=\frac{1}{n^2}\e{\sum_{i,j}\left(\x_j^\intercal \x_i\right)} =\frac{1}{n^2}\sum_{i,j}\e{\left(\x_j^\intercal \x_i\right)}  = \frac{1}{n^2}\left(\sum_{i=1}^n\e{(\x_i^\intercal \x_i)}+\sum_{i=1}^n\sum_{j\not=i}^n \e{(\x_i^\intercal \x_j)}\right)\\ =\frac{1}{n^2} \sum_{i=1}^n \e{\norm{\x_i}^2}\leq \frac{1}{n}\sigma^2,
\end{aligned}
\end{equation}
where we used the fact that the expectation of a sum equals the sum of the expectations and the cross-terms $\e{\x_j^\intercal\x_i}=0, j\not=i$ because of the independence assumption. Hence, we can bound the term $V\leq \frac{1}{n}\sigma^2$ for the random vector sum $\z$. 

Now, since $n>1$ and $\epsilon >0$, as well as $P(\z>a)$ is falling in $a$ and $\exp(-x)$ falling in $x$, we can use this upper bound on the variance of $\z$ in (\ref{eq:modified_vector_bernstein}), which gives the desired inequality

\begin{equation}
P\left(\norm{\z}\geq \epsilon\right) \leq \exp\left(-n\cdot \frac{\epsilon^2}{8\sigma^2}+\frac{1}{4}\right)
\end{equation}
\begin{flushright}
$\square$
\end{flushright}

This result was applied in order to find the probabilistic bound on the deviation of the sub-sampled gradient from the full gradient as stated in Lemma \ref{l:gradient_deviation}, for which we will give the proof next.

\textbf{\textit{Proof of Lemma \ref{l:gradient_deviation}:}}

To apply vector Bernstein's inequality (\ref{eq:final_bernstein_vector}) we need to center the gradients. Thus we define
\begin{equation}
\x_i=g_i(\x)-\nabla f(\x),\: i=1,\ldots,|S_g|
\end{equation}
and note that from the Lipschitz continuity of $f$ (A\ref{a:continuity}), we have
\begin{equation}
\norm{\x_i}=\norm{\g_i(\x)-\nabla f(\x)}\leq \norm{\g_i(\x)}+\norm{\nabla f(\x)}\leq 2\kappa_f \text{ and } \norm{\x_i}^2 \leq 4\kappa_f^2,\: i=1,\ldots,|S_g|.
\end{equation}
With $\sigma^2 := 4\kappa_f^2$ and 
\begin{equation}
\z=\frac{1}{|S_g|}\sum_{i \in S_g} \x_i=\frac{1}{|S_g|}\sum_{i \in S_g}\g_i(\x)-\frac{1}{|S_g|}\sum_{i \in S_g}\nabla f(\x) = \g(\x)-\nabla f(\x)
\end{equation}
in equation (\ref{eq:final_bernstein_vector}), we can require the probability of a deviation larger or equal to $\epsilon$ to be lower than some $\delta \in (0,1]$
\begin{equation}\label{temp:gradient_deviations}
\begin{aligned}
P\left(\norm{\g(\x)-\nabla f(\x)}>\epsilon \right)\leq & 2d\exp \left(-|S_g|\cdot \frac{\epsilon^2}{32\kappa_f^2}+\frac{1}{4}\right) \overset{!}{\leq} \delta \\
\Leftrightarrow & |S_g|\cdot \frac{\epsilon^2}{32\kappa_f^2}-\frac{1}{4} \overset{!}{\geq} \log((2d)/\delta)\\
\Leftrightarrow & \epsilon \geq 4\sqrt{2}\kappa_f\sqrt{\frac{\log \left((2d)/\delta\right)+1/4}{|S_g|}}.
\end{aligned}
\end{equation}
Conversely, the probability of a deviation of
\begin{equation}
\epsilon < 4\sqrt{2}\kappa_f\sqrt{\frac{\log \left((2d)/\delta\right)+1/4}{|S_g|}}
\end{equation}
is higher or equal to $1-\delta$.
\begin{flushright}
$\square$
\end{flushright}

Of course, any sampling scheme that guarantees the right hand side of (\ref{eq:g_deviation}) to be smaller or equal to $M$ times the squared step size, directly satisfies the sufficient gradient agreement condition (A\ref{a:Strong_agreement_g}). Consequently, plugging the former into the latter and rearranging for the sample size gives Theorem \ref{t:grad_sampling} as we shall prove now.

\textbf{\textit{Proof of Theorem \ref{t:grad_sampling}:}}

By use of Lemma \ref{l:gradient_deviation} we can write
\begin{equation}
\begin{aligned}
&\: \norm{\g(\x)-\nabla f(\x)} \leq M\norm{\s}^2 \\
\Leftrightarrow &\: 4\sqrt{2}\kappa_f \sqrt{\frac{\log(1/\delta+1/4)}{|S_{g}|}} \leq M \norm{\s}^2 \\
&\: |S_{g}| \geq \dfrac{32\kappa_f^2\log\left(1/\delta+1/4\right)}{M^2\norm{\s}^4}
\end{aligned}
\end{equation}
\begin{flushright}
$\square$
\end{flushright}

\newpage
\subsubsection{Hessian Sampling}
\begin{framed}
\begin{lemma}[Matrix Bernstein Inequality]
Let $\Am_1,..,\Am_n$ be independent random Hermitian matrices with common dimension $d \times d$ and assume that each one is centered, uniformly bounded and also the variance is bounded above:
\begin{equation*}
\e{\Am_i}=0 \text{ and } \norm{\Am_i}_2 \leq \mu \text{ as well as } \norm{\e{\Am_i^2}}_2 \leq \sigma^2
\end{equation*}
Introduce the sum 
\begin{equation*}
\Zm=\frac{1}{n}\sum_{i=1}^n \Am_i
\end{equation*}
Then we have
\begin{equation} \label{eq:bernstein2}
P(\norm{\Zm}\geq \epsilon) \leq 2d \cdot \exp\left(-n\cdot \min \{\frac{\epsilon^2}{4\sigma^2},\frac{\epsilon}{2\mu} \}\right)
\end{equation}

\end{lemma}
\end{framed}
\textit{Proof:}
Theorem 12 in \cite{gross2011recovering} gives the following Operator-Bernstein inequality
\begin{equation}
P \left( \norm{\sum_{i=1}^n \Am_i}\geq \epsilon \right) \leq 2d \cdot \exp\left(\min \{\frac{\epsilon^2}{4V},\frac{\epsilon}{2\mu} \}\right),
\end{equation}
where $V=n\sigma^2$. As well shall see, this is an upper bound on the variance of $\Ym=\sum_{i=1}^n \Am_i$ since the $\Am_i$ are independent and have an expectation of zero ($\e{Y}=0$).
\begin{equation}
\begin{aligned}
Var(\Ym)=& \norm{\e{\Ym^2}-\e{\Ym}^2}=\norm{\e{(\sum_{i} \Am_i)^2}}=\norm{\e{\sum_{i,j} \Am_i\Am_j}}  =\norm{\sum_{i,j}\e{\Am_i\Am_j}}\\
=&\norm{\sum_{i}\e{\Am_i\Am_i} + \sum_{i}\sum_{j\not=i}\e{\Am_i\Am_j} }=\norm{\sum_i \e{\Am_i^2}}\leq \sum_i \norm{\e{\Am_i^2}}\leq n\sigma^2,
\end{aligned}
\end{equation}
where we used the fact that the expectation of a sum equals the sum of the expectations and the cross-terms $\e{\Am_j\Am_i}=0, j\not=i$ because of the independence assumption.

However, $\Zm=\frac{1}{n}\sum_{i=1}^n \Am_i$ and thus
\begin{equation}\label{temp:bernstein1}
Var(\Zm)= \norm{\e{\Zm^2}}=\norm{\e{(\frac{1}{n}\sum_{i=1}^n \Am_i)^2}}=\frac{1}{n^2}\norm{\e{(\sum_{i=1}^n \Am_i)^2}}\leq \frac{1}{n} \sigma^2.
\end{equation}
Hence, we can bound $V\leq \frac{1}{n}\sigma^2$ for the \textit{average} random matrix sum $\Zm$. Furthermore, since $n>1$ and $\epsilon, \mu >0$ as well as $\exp(-\alpha)$ decreasing in $\alpha \in \mathbb{R}$ we have that
\begin{equation}\label{temp:bernstein2}
\exp\left(-\frac{\epsilon}{2\mu}\right) \leq \exp\left(-\frac{\epsilon}{n2\mu}\right).
\end{equation}

Together with the Operator-Bernstein inequality, (\ref{temp:bernstein1}) and (\ref{temp:bernstein2}) give the desired inequality (\ref{eq:bernstein2}).
\begin{flushright}
$\square$
\end{flushright}

This result exhibits that sums of independent random matrices provide normal concentration near its mean in a range determined by the variance of the sum. We  apply it in order to derive the bound on the deviation of the sub-sampled Hessian from the full Hessian as stated in Lemma \ref{l:Hessian_deviation}, which we shall prove next.

\textbf{\textit{Proof of Lemma \ref{l:Hessian_deviation}:}}
Bernstein's Inequality holds as $f\in C^2$ and thus the Hessian is symmetric by Schwarz's Theorem. Since the expectation of the random matrix needs to be zero, we center the individual Hessians,
\begin{equation*}
\Xm_i=\Hm_i(\x)-\Hm(\x), i=1,...,|S_H|
\end{equation*} and note that now from the Lipschitz continuity of $\g$ (A\ref{a:continuity}):
\begin{equation*}
\norm{\Xm_i}_2 \leq 2\kappa_g , i=1...|S_H| \text{ and } \norm{\Xm_i^2}_2 \leq 4\kappa_g^2 , i=1...|S_H| .
\end{equation*}
Hence, for $\epsilon\leq 4\kappa_g$, we are in the \textit{small deviation} regime of Bernstein's bound with a sub-gaussian tail. Then, we may plug
\begin{equation*}
\frac{1}{|S_H|} \sum_{i=1}^{|S_H|} \Xm_i = \Bm(\x)-\Hm(\x) 
\end{equation*}
into (\ref{eq:bernstein2}), to get
\begin{equation}
\:P(\norm{\Bm(\x)-\Hm(\x)}\geq \epsilon ) \leq 2d\cdot \exp\left(-\frac{\epsilon^2|S_{H}|}{16\kappa_g^2}\right).\:
\end{equation}

Finally, we shall require the probability of a deviation of $\epsilon$ or higher to be lower than some $\delta \in (0,1]$
\begin{equation}
\begin{aligned}
&\: 2d\cdot \exp\left(-\frac{\epsilon^2|S_{H}|}{16\kappa_g^2}\right) \overset{!}{=} \delta \\
\Leftrightarrow &\: -\frac{\epsilon^2|S_{H}|}{16\kappa_g^2}=\log(\delta/2d) \\
\Leftrightarrow &\:  \epsilon = 4\kappa_g \sqrt{\frac{\log(2d/\delta)}{|S_{H}|}},
\end{aligned}
\end{equation}
which is equivalent to $\norm{\Bm(\x)-\Hm(\x)}$ staying within this particular choice of $\epsilon$ with probability $(1-\delta)$, generally perceived as \textit{high probability}.
\begin{flushright}
$\square$
\end{flushright}

\textbf{\textit{Proof of Theorem \ref{t:hessian_sampling}:}}
Since $\|\Am\vb\| \le \|\Am\|_{op} \|\vb\| \:\mbox{ for every } \vb \in V$ we have for the choice of the spectral matrix norm and euclidean vector norm that any $\Bm$ that fulfils  $\norm{(\Bm(\x)-\Hm(\x))} \leq C\norm{\s}$ also satisfies condition A\ref{a:Strong_agreement_H}. Furthermore
\begin{equation}\label{eq:samplesize}
\begin{aligned}
&\norm{(\Bm-\Hm(\x))} \leq C\norm{\s}\\
\Leftrightarrow &\: 4\kappa_g\sqrt{\frac{\log(2d/\delta)}{|S_{H}|}}
 \leq C\norm{\s} \\
\Leftrightarrow &\: |S_{H}| \geq \frac{16\kappa_g^2\log(2d/\delta)}{(C\norm{\s})^2},&\: C>0.
\end{aligned}
\end{equation}
\begin{flushright}
$\square$
\end{flushright}

Note that there may be a less restrictive sampling conditions that satisfy A\ref{a:Strong_agreement_H} since condition (\ref{eq:samplesize}) is based on the worst case bound $\|\Am\vb\| \le \|\Am\|_{op} \|\vb\|$ which indeed only holds with equality if $\vb$ happens to be (exactly in the direction of) the largest eigenvector of $A$.

Finally, we shall state a Lemma which illustrates that the stepsize goes to zero as the algorithm converges. The proof can be found in Section 5 of \citep{cartis2011adaptive}. 
\begin{framed}
\begin{lemma}\label{l:s_to_0}
Let $\{f(\x_k)\}$ be bounded below by some $f_{\inf}>-\infty$. Also, let $\s_k$ satisfy A\ref{a:approx_model_min} and $\sigma_k$ be bounded below by some $\sigma_{\inf}>0$. Then we have for all successful iterations that
\begin{equation}\label{eq:s_to_0}
\norm{\s_k} \rightarrow 0,\text{ as } k\rightarrow \infty
\end{equation}
\end{lemma}
\end{framed}
\subsubsection{Illustration}

\begin{figure*}[h!]
	\begin{center}
          \begin{tabular}{@{}c@{\hspace{0.5mm}}c@{\hspace{0.5mm}}c@{}}
            \includegraphics[width=0.33\linewidth]{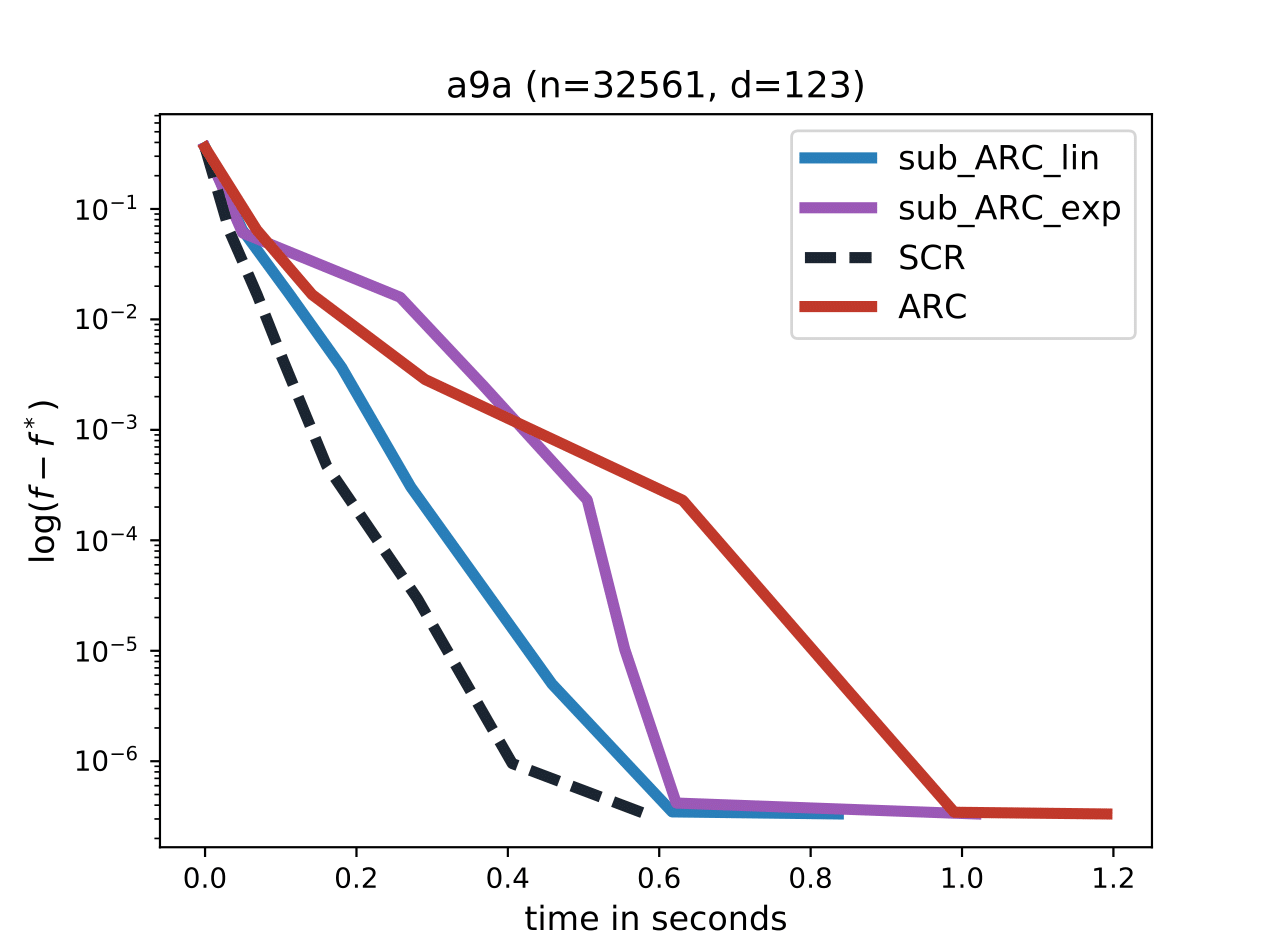} &
            \includegraphics[width=0.33\linewidth]{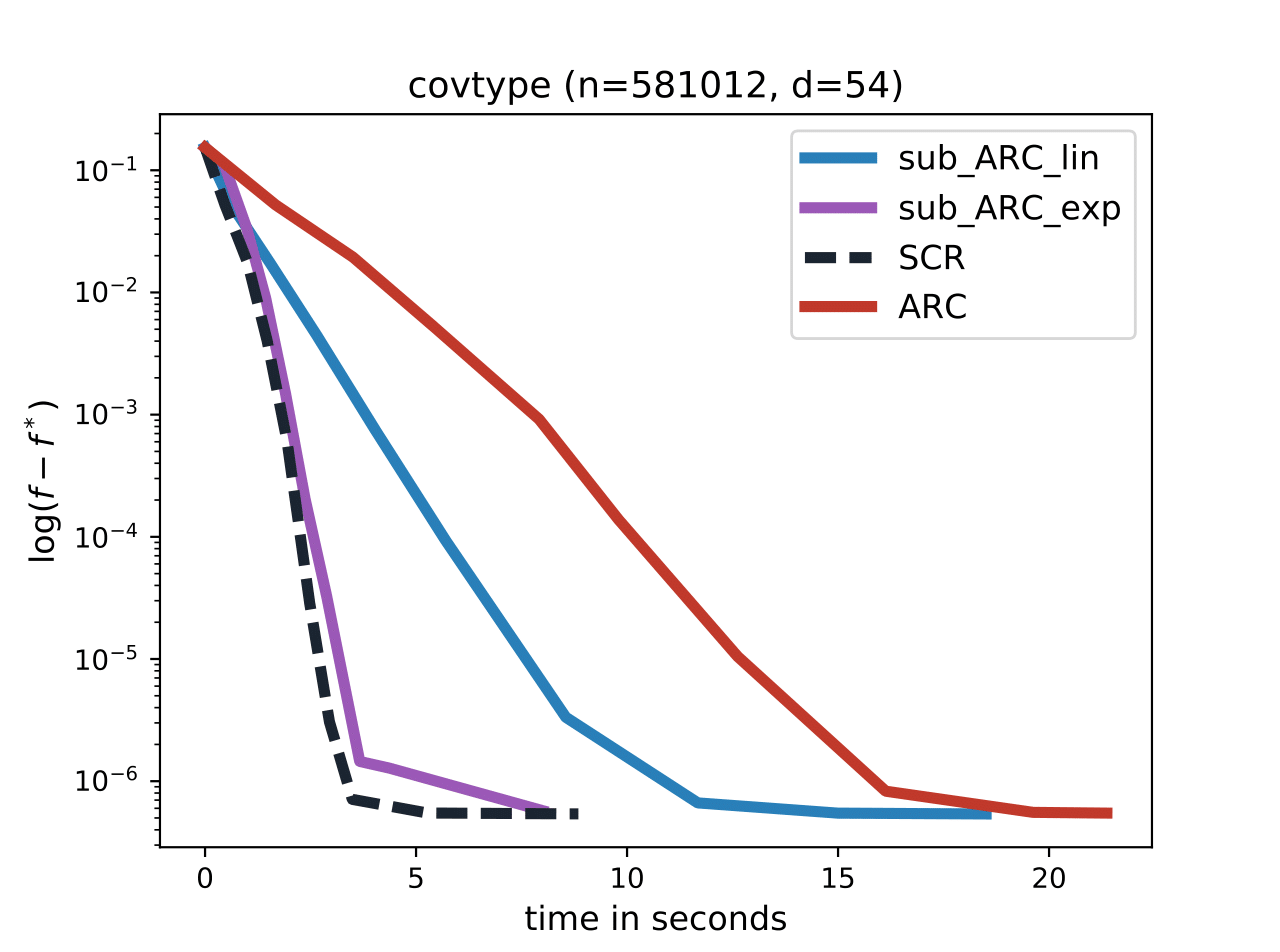}  &
             \includegraphics[width=0.33\linewidth]{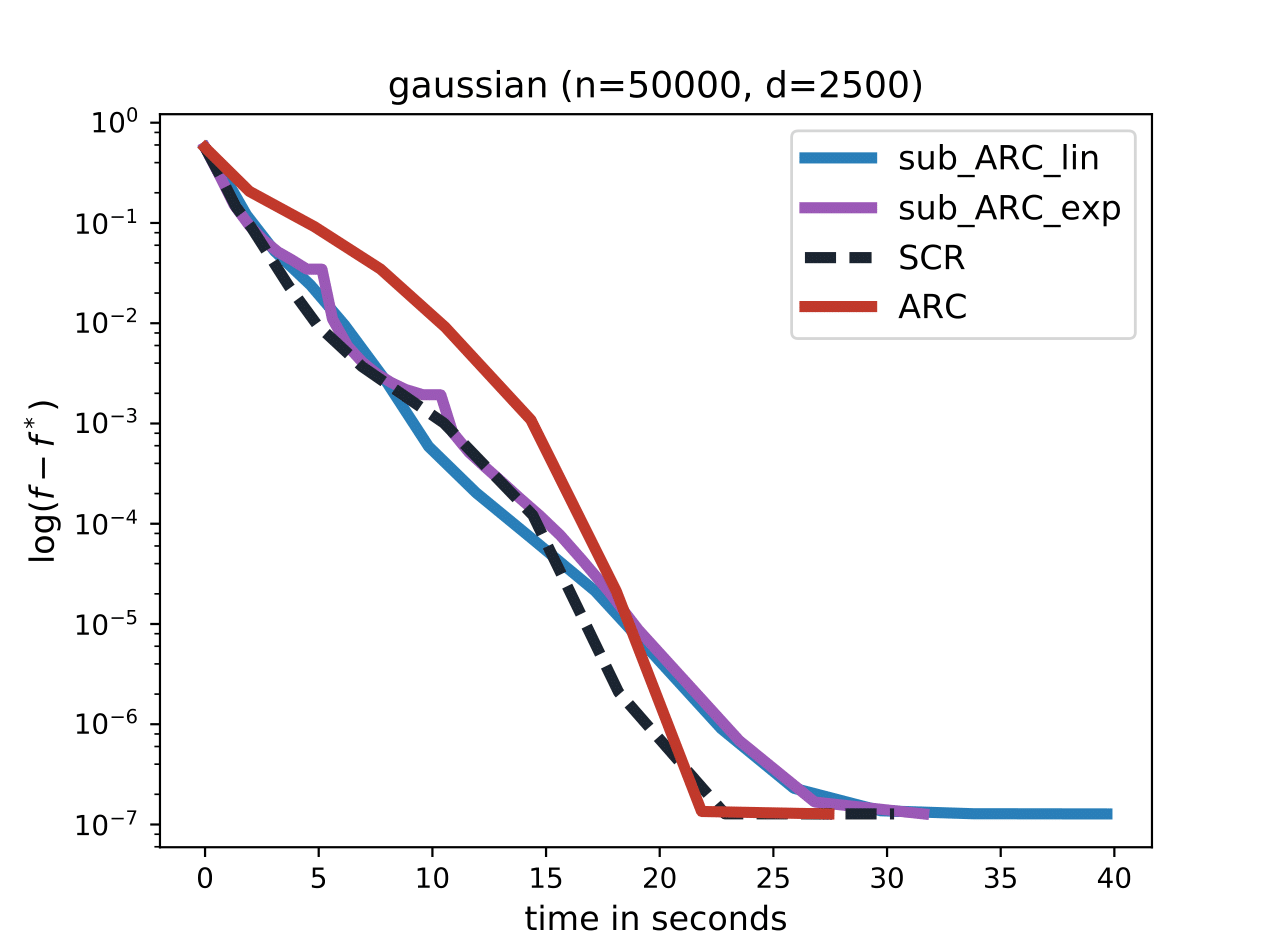}    \\ 
             \includegraphics[width=0.33\linewidth]{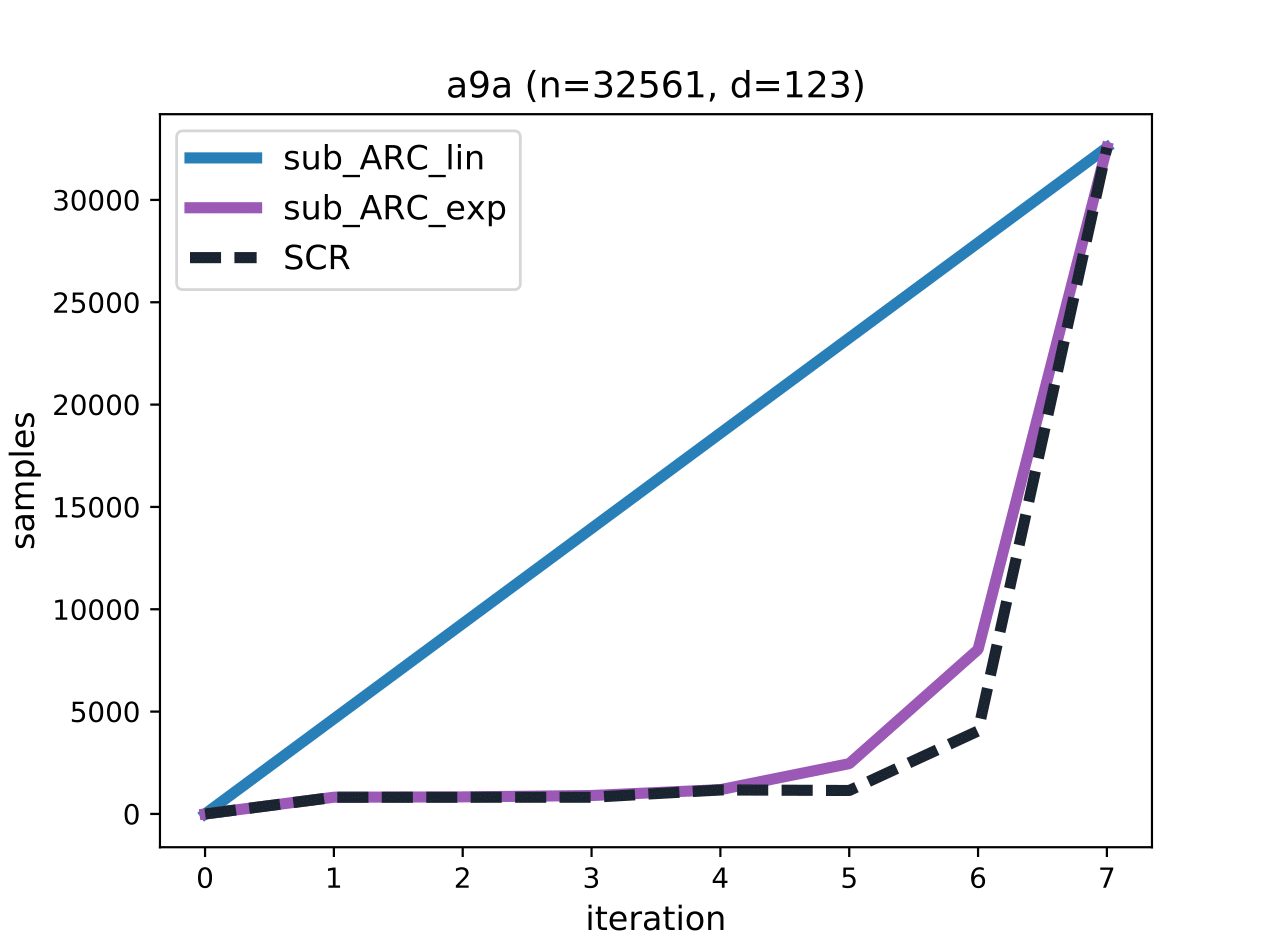} &
            \includegraphics[width=0.33\linewidth]{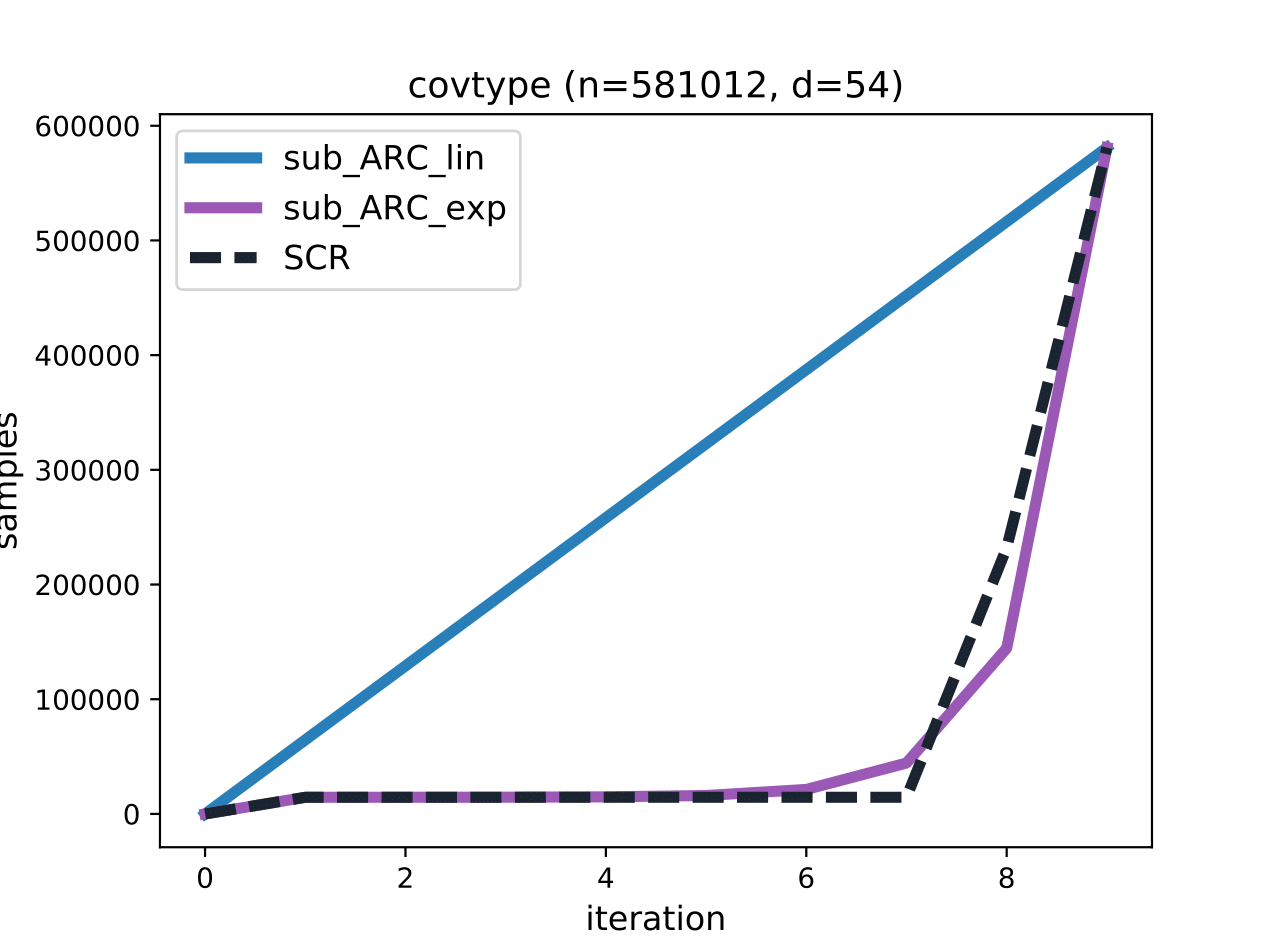}  &
              \includegraphics[width=0.33\linewidth]{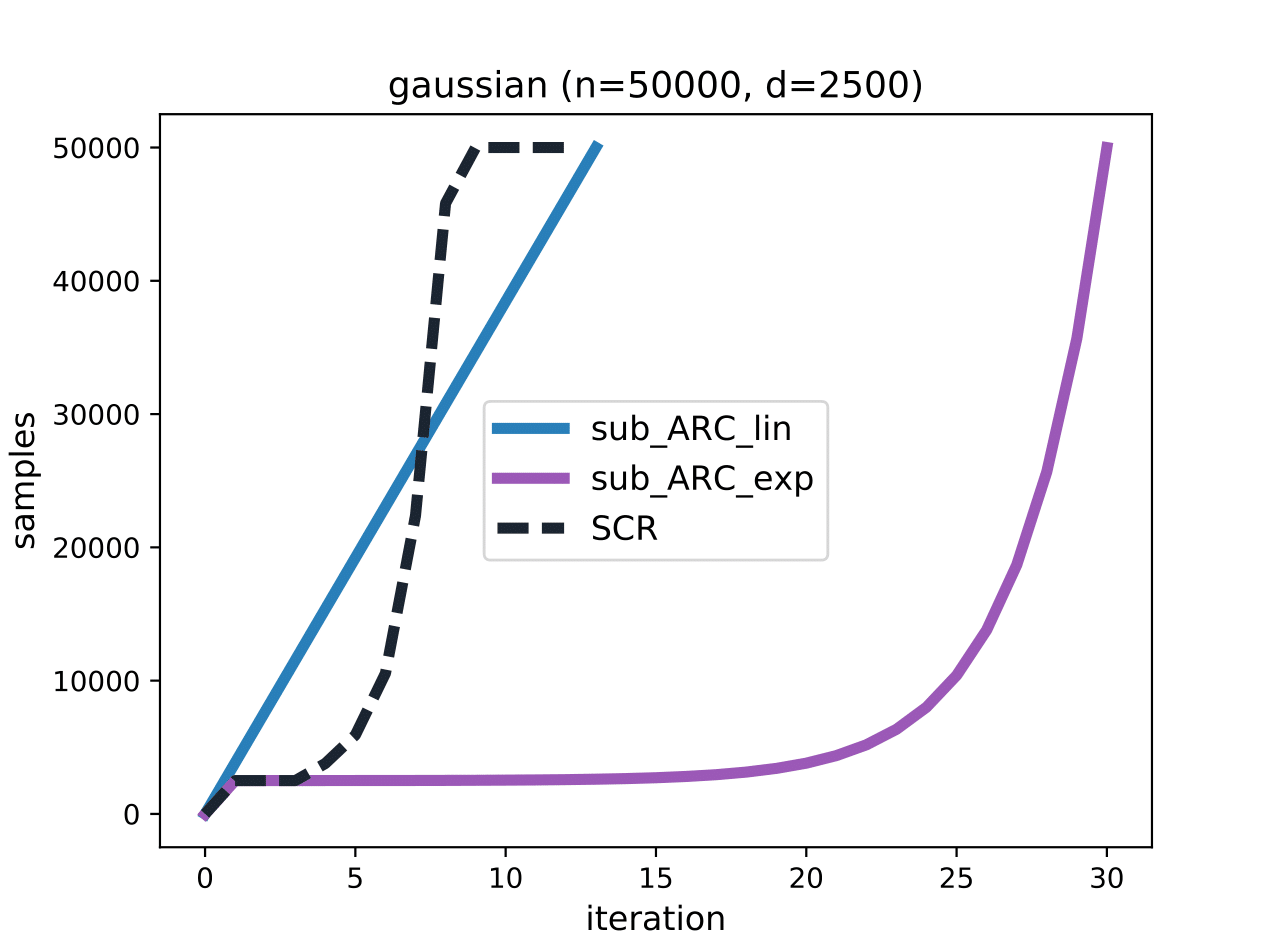}     \\ 
         	1. {\sc a9a} &
            2. {\sc covtype} &
            3. {\sc gaussian} 
	  	\end{tabular}
          \caption{Suboptimality (top row) and sample sizes (bottom row) for different cubic regularization methods on \textit{a9a}, \textit{covtype} and \textit{gaussian}. Note that the automatic sampling scheme of SCR follows an exponential curve, which means that it can indeed save a lot of computation in the early stage of the optimization process.}
          \label{fig:sample_sizes}
	\end{center}
\end{figure*}

In the top row of Figure~\ref{fig:sample_sizes} we illustrate the Hessian sample sizes that result when applying SCR with a practical version of Theorem \ref{t:hessian_sampling} to the datasets used in our experiments \footnote{see Section \ref{sec:exp_details} for details}. In the bottom row of Figure~\ref{fig:sample_sizes}, we benchmark our algorithm to the deterministic as well as two naive stochastic versions of ARC with \textit{linearly} and \textit{exponentially} increasing sample sizes. 

Note that both the linear and the exponential sampling schemes do not quite reach the same performance as SCR even though they were carefully fine tuned to achieve the best possible performance. Furthermore, the sampling size was manually set to reach the full sample size at the very last iteration. This highlights another advantage of the automatic sampling scheme that does not require knowledge of the total number of iterations.

\subsection{Convergence Analysis}
\subsubsection{Preliminary results}
\textbf{\textit{Proof of Lemma \ref{l:bounds_on_sigma}:}}

The lower bound $\sigma_{\inf}$ follows directly from Step 7 in the algorithm design (see Algorithm \ref{alg:src}). Within the upper bound, the constant $\sigma_0$ accounts for the start value of the penalty parameter. Now, we show that as soon as some $\sigma_k>3(\frac{2M+C+\kappa_g}{2})$, the iteration is very successful and $\sigma_{k+1}<\sigma_k$. Finally, $\gamma_2$ allows for $\sigma_k$ being 'close to' the successful threshold, but increased 'one last time'. 

Any iteration with $f(\x_k+\s_k)\leq m(\s_k)$ yields a $\rho_k\geq1\geq \eta_2$ and is thus very successful. From a 2nd-order Taylor approximation of $f(\x_k+\s_k)$ around $\x_k$ we have:
\begin{equation}
\begin{aligned}
f(\x_k+\s_k)-m_k(\s_k)=&\: (\nabla f(\x_k) - \g(\x_k))^\intercal \s_k+\frac{1}{2} \s_k^\intercal (\Hm(\x_k+t\s_k)-\Bm_k) \s_k - \frac{\sigma}{3}\norm{\s_k}^3\\
\leq &\: \evec_k^\intercal \s_k + \frac{1}{2}\norm{\s_k}^2\norm{\Hm(\x_k+t\s_k)-\Hm(\x)}+\frac{1}{2}\norm{\Hm(\x_k)-\Bm_k}\norm{\s_k} - \frac{\sigma_k}{3}\norm{\s_k}^3\\
\leq & \: \norm{\bf{e}_k}\norm{\s_k}+ \left(\frac{C+\kappa_g}{2}-\frac{\sigma_k}{3}\right)\norm{\s_k}^3\\
\leq & \: M\norm{\s_k}^3 + \left(\frac{C+\kappa_g}{2}-\frac{\sigma_k}{3}\right)\norm{\s_k}^3\\=&\left(\frac{2M+C+\kappa_g}{2}-\frac{\sigma_k}{3}\right)\norm{\s_k}^3
\end{aligned}
\end{equation}

Requiring the right hand side to be non-positive and solving for $\sigma_k$ gives the desired result.\begin{flushright}
$\square$
\end{flushright}

\textbf{\textit{Proof of Lemma \ref{l:lower_bound_function_decrease} :}}
By definition of the stochastic model $m_k(\s_k)$ we have
\begin{equation}
\begin{aligned}
f(\x_k)-m_k(\s_k) =& - \s_k^\intercal \g(\x_k) -\frac{1}{2}\s_k^\intercal \Bm_k \s_k-\frac{1}{3}\sigma_k \norm{\s_k}^3\\
=  & \frac{1}{2} \s_k^\intercal \Bm_k \s_k +\frac{2}{3}\sigma_k\norm{\s_k}^3 \\
\geq & \frac{1}{6} \sigma_k\norm{\s_k}^3,
\end{aligned}
\end{equation}
where we applied equation (\ref{eq:3.11}) first and equation (\ref{eq:3.12}) secondly. \begin{flushright}
$\square$
\end{flushright}

Before proving the lower bound on the stepsize $\|\s_k\|$ we first transfer the rather technical result from Lemma 4.6 in \citep{cartis2011adaptive} to our framework of stochastic gradients. For this purpose, let $\eb_k$ be the gradient approximation error, i.e. $\eb_k:=\g_k - \nabla f(\x_k)$. 

\begin{framed}
\begin{lemma}\label{l:technical_result}
Let $f\in C^2$, Lipschitz continuous gradients (A\ref{a:continuity}) and TC (A\ref{a:TC}) hold. Then, for each (very-)successful $k$, we have
\begin{equation}\label{eq:L4.6} 
\begin{aligned}
&(1-\kappa_\theta) \norm{\nabla f(\x_{k+1})} \:\leq \sigma_k\norm{\s_k}^2 + \\ & \underbrace{\left( \norm{\int_0^1(\Hm(\x_k+t\s_k)-\Hm(\x_k))dt} + \dfrac{\norm{(\Hm(\x_k)-\Bm_k)\s_k}}{\norm{\s_k}} +\kappa_\theta\kappa_g\norm{\s_k} +  (1+\kappa_\theta\kappa_g)\dfrac{\norm{\eb_k}}{\norm{\s_k}} \right)}_{=d_k} \cdot \norm{\s_k}
\end{aligned}
\end{equation}
with $\kappa_\theta \in (0,1)$ as in TC (\ref{eq:TC}).
\end{lemma}
\end{framed}
\textit{Proof:} We shall start by writing

\begin{equation}
\begin{aligned}
\norm{\nabla f(\x_k+\s_k)} \leq &\: \norm{ \nabla f(\x_k+\s_k)-\nabla m_k(\s_k)}+\norm{\nabla m_k(\s_k)} \\
\leq & \: \underbrace{\norm{ \nabla f(\x_k+\s_k)-\nabla m_k(\s_k)}}_{(a)} +\underbrace{\theta_k\norm{\g_k(\x_k)}}_{(b)}, 
\end{aligned}
\end{equation}
where the last inequality results from TC (Eq.~(\ref{eq:TC})). 
Now, we can find the following bounds on the individual terms:

\textbf{(a)} By (\ref{eq.subproblem_abl}) we have 
\begin{equation}\label{eq:temp_a}
\norm{\nabla f(\x_k+\s_k)-\nabla m_k} = \norm{\nabla f(\x_k+\s_k)-\g_k(\x_k)-\Bm_k \s_k-\sigma_k\s_k\norm{\s_k}}.
\end{equation}
We can rewrite the right-hand side by a Taylor expansion of $\nabla f_{k+1}(\x_k+\s_k)$ around $\x_k$ to get
\begin{equation}
(\ref{eq:temp_a}) = \norm{\nabla f(\x_k)+\int_0^1 \Hm(\x_k+t\s_k)\s_kdt -\g_k(\x_k)- \Bm_k\s_k-\sigma_k\s_k\norm{\s_k}}.
\end{equation}
Contrary to the case of deterministic gradients, the first and third summand no longer cancel out. Applying the triangle inequality repeatedly, we thus get an error term in the final bound on (a):
\begin{equation}
\begin{aligned}
\norm{\nabla f(\x_k+\s_k)-\nabla m_k}\leq & \: \norm{\int_0^1 \Hm((\x_k+t\s_k)-\Bm_k) \s_kdt} +\sigma_k\norm{\s_k}^2 + \norm{\nabla f(\x_k)-\g_k(\x_k)} \\
\leq &\: \norm{\int_0^1 \Hm((\x_k+t\s_k)dt - \Hm(\x_k)}\cdot \norm{\s_k} + \norm{(\Hm(\x_k)-\Bm_k)\s_k}\\ &\: +\sigma_k\norm{\s_k}^2 + \norm{\eb_k}.
\end{aligned}
\end{equation}

\textbf{(b)} To bound the second summand, we can write
\begin{equation}
\begin{aligned}
\norm{\g(\x_k)}\leq & \: \norm{\nabla f(\x_k)}+\norm{\eb_k} \\
\leq & \: \norm{\nabla f(\x_k+\s_k)} + \norm{\nabla f(\x_k)-\nabla f(\x_k+\s_k)} +\norm{\eb_k}\\
\leq & \: \norm{\nabla f(\x_k+\s_k)} + \kappa_g\norm{\s_k} + \norm{\eb_k}
\end{aligned}
\end{equation}

Finally, using the definition of $\theta_k$ as in (\ref{eq:TC}) (which also gives $\theta_k\leq\kappa_\theta$ and $\theta_k\leq\kappa_\theta h_k$) and combining (a) and (b) we get the above result. \begin{flushright}
$\square$
\end{flushright}

\textbf{\textit{Proof of Lemma \ref{l:lower_bound_on_s}:}}
The conditions of Lemma \ref{l:technical_result} are satisfied. By multiplying $d_k\norm{\s_k}$ out in equation (\ref{eq:L4.6}), we get
\begin{equation}
\begin{aligned}
& (1-\kappa_\theta)\norm{\nabla f(\x_{k+1})} \leq \\ &  \norm{\int_0^1(\Hm(\x_k+t\s_k)-\Hm(\x_k))dt} \norm{\s_k} + \norm{(\Hm(\x_k)-\Bm_k)\s_k} +\kappa_\theta\kappa_g\norm{\s_k}^2 + (1+\kappa_\theta\kappa_g) \norm{\eb_k} + \sigma_k\norm{\s_k}^2.
\end{aligned}
\end{equation}

Now, applying the strong agreement conditions (A\ref{a:Strong_agreement_H}) and (A\ref{a:Strong_agreement_g}), as well as the Lipschitz continuity of H, we can rewrite this as

\begin{equation}
(1-\kappa_\theta)\norm{\nabla f(\x_{k+1})} \leq (\frac{1}{2}\kappa_g+C+(1+\kappa_\theta\kappa_g)M+\sigma_{\max}+\kappa_\theta\kappa_g) \norm{\s_k}^2 ,
\end{equation}
for all sufficiently large, successful $k$. Solving for the stepsize $\norm{\s_k}$ give the desired result.\begin{flushright}
$\square$
\end{flushright}

\subsubsection{Local convergence}
Before we can study the convergence rate of SCR in a locally convex neighbourhood of a local minimizer $w_*$ we first need to establish three crucial properties:
\begin{enumerate}
\item a lower bound on $\norm{\s_k}$ that depends on $\norm{\g_k}$.
\item an upper bound on $\norm{\s_k}$ that depends on $\norm{\g_{k+1}}$.
\item an eventually full sample size
\item conditions under which all steps are eventually very successful.
\end{enumerate}
With this at hand we will be able to relate $\norm{\g_{k+1}}$ to $\norm{\g_k}$, show that this ratio eventually goes to zero at a quadratic rate and conclude from a Taylor expansion around $\g_k$ that the iterates themselves converge as well. 

\begin{assumption}[Sampling Scheme]\label{a:sampling_scheme}
Let $\g_k$ and $\Bm_k$ be sampled such that \ref{eq:gradient_sampling_scheme} and \ref{eq:hessian_sampling_scheme} hold in each iteration $k$. Furthermore, for unsuccessful iterations, assume that the sample size is not decreasing.
\end{assumption}

We have already established a lower stepsize bound in Lemma \ref{l:lower_bound_on_s} so let us turn our attention directly to 2.:
\begin{lemma}[Upper bound on stepsize]\label{l:upper_bound_on_s}
Suppose that $\s_k$ satisfies (\ref{eq:3.11}) and that the Rayleigh coefficient
\begin{equation}
R_k(\s_k) := \dfrac{\s_k^\intercal \Bm_k\s_k}{\norm{\s_k}^2}
\end{equation}
is positive, then
\begin{equation}\label{eq:upper_bound_on_s}
\norm{\s_k}\leq \dfrac{1}{R_k(\s_k)} \norm{\g_k}=\dfrac{1}{R_k(\s_k)} \norm{\nabla f(\w_k) + \evec_k} \leq \dfrac{1}{R_k(\s_k)} (\norm{\nabla f(\w_k)} + \norm{\evec_k})
\end{equation}
\end{lemma}
\textit{Proof:}
Given the above assumptions we can rewrite (\ref{eq:3.11}) as follows
\begin{equation}\label{temp:3.11}
R_k(\s_k)\norm{\s_k}^2 = -\s_k^\intercal \g_k - \sigma_k \norm{\s_k}^3 \leq \norm{\s_k}\norm{\g_k},
\end{equation}
where we used Cauchy-Schwarz inequality as well as the fact that $\sigma_k >0,\:\forall k$. Solving (\ref{temp:3.11}) for $\norm{\s_k}$ gives (\ref{eq:upper_bound_on_s}).
\begin{flushright}
$\square$
\end{flushright}

\begin{lemma}[Eventually full sample size]\label{l:eventu_full_samples}
Let $\{f(\x_k)\}$ be bounded below by some $f_{\inf}>-\infty$. Also, let A\ref{a:approx_model_min}, A\ref{a:continuity} hold and let $\g_k$ and $\Bm_k$ be sampled according to A\ref{a:sampling_scheme}. Then we have w.h.p. that
\begin{equation}
|S_{g,k}| \rightarrow n \text{ and } |S_{B,k}| \rightarrow n \text{ as } k \rightarrow \infty
\end{equation}
\end{lemma}

The sampling schemes from Theorem \ref{t:grad_sampling} and Theorem \ref{t:hessian_sampling} imply that the sufficient agreement assumptions A\ref{a:Strong_agreement_g} and A\ref{a:Strong_agreement_H} hold with high probability. Thus, we can deduce from Lemma 10 that after a certain number of consecutive unsuccessful iterates the penalty parameter is so high ($\sigma_k \geq \sigma_{sup}$) that we are guaranteed to find a successful step. Consequently, the number of successful iterations must be infinite ($|\mathcal{S}|=\infty$) when we consider the asymptotic convergence properties of SCR. We are left with two possible scenarios:

(i) If the number of unsuccessful iterations is finite ($|\mathcal{U}|\leq \infty$ \& $|\mathcal{S}|=\infty$) we have that $\exists\; \hat{k}$ after which all  iterates are successful, i.e. $k \in \mathcal{S}, \forall \; k>\hat{k}$. From Lemma 20 we know that for all successful iterations $\norm{\s_k}\rightarrow 0 \text{ as } k\rightarrow\infty$. Consequently, due to the sampling scheme as specified in Theorem \ref{t:grad_sampling} and Theorem \ref{t:hessian_sampling}, $\exists\; \bar{k} \geq \hat{k}$ with $|S_{g,k}| =|S_{B,k}| = n, \;\forall\; k\geq \bar{k}$. 

(ii) If the number of unsuccessful iterations is infinite ($|\mathcal{U}|=\infty$ \& $|\mathcal{S}|=\infty$) we know for the same reasons that for the \textit{sub}sequence of successful iterates $\{k=0,1,\ldots \infty|k\in \mathcal{S}\}$ again $\norm{\s_k}\rightarrow0 $, as $ k\in \mathcal{S} \rightarrow\infty$ and hence $\exists\; \tilde{k}$ with $|S_{g,k}|= |S_{B,k}| = n, \;\forall\; k\geq \tilde{k}\in \mathcal{S}$. Given that we do specifically not decrease the sample size in unsuccessful iterations we have that  $|S_{g,k}|= |S_{B,k}| =n,\; \forall\; k\geq \tilde{k}$.

As a result the sample sizes eventually equal $n$ with high probability in all conceivable scenarios which proves the assertion\footnote{We shall see that, as a result of Lemma \ref{l:eventu_succ_iter}, the case of an infinite number of unsuccessful steps can actually not happen}.
\begin{flushright}
$\square$
\end{flushright}

Now that we have (asymptotic) stepsize bounds and gradient (Hessian) agreement we are going to establish that, when converging, all SCR iterations are indeed very successful asymptotically. 
\begin{lemma}[Eventually successful iterations]\label{l:eventu_succ_iter}
Let $f\in C^2$, $\nabla f$ uniformly continuous and $\Bm_k$ bounded above. Let $\Bm_k$ and $\g_k$ be sampled according to A\ref{a:sampling_scheme}, as well as $\s_k$ satisfy (\ref{eq:3.11}). Furthermore, let
\begin{equation}\label{temp:limit1}
\w_k\rightarrow \w_*,\: \text{as } k \rightarrow \infty,
\end{equation}
with $\nabla f(\w_*)=0$ and $\Hm(\w_*)$ positive definite. Then there exists a constant $R_{min}>0$ such that for all $k$ sufficiently large
\begin{equation} \label{temp:R}
R_k(\s_k) \geq R_{\min}.
\end{equation}
Furthermore, all iterations are eventually very successful w.h.p.
\end{lemma}
\textit{Proof:}
Since $f$ is continuous, the limit (\ref{temp:limit1}) implies that $\left\{f(\w_k)\right\}$ is bounded below. Since $\Hm(\w_*)$ is positive definite per assumption, so is $\Hm(\w_k)$ for all $k$ sufficiently large. Therefore, there exists a constant $R_{\min}$ such that
\begin{equation}
\frac{\s_k^\intercal \Hm(\w_k)\s_k}{\norm{\s_k}^2} >2R_{\min} > 0, \text{	for all } k \text{ sufficiently large.}
\end{equation} 

As a result of Lemma \ref{l:eventu_full_samples} we have that $\norm{\evec_k}\rightarrow 0$ as $k \rightarrow \infty$. Hence, Lemma \ref{l:upper_bound_on_s} yields $\norm{\s_k} \leq 1/R_{\min} \norm{\nabla f_k}$ which implies that the step size converges to zero as we approximate $w^*$. Consequently, we are able to show that eventually all iterations are indeed very successful. Towards this end we need to ensure that the following quantity $r_k$ becomes negative for sufficiently large $k$:
\begin{equation}
r_k:=\underbrace{f(\w_k+\s_k)-m(\s_k)}_{(i)} +(1-\eta_2)\underbrace{(m(\s_k)-f(\w_k))}_{(ii)},
\end{equation}
where $\eta_2 \in (0,1)$ is the "very successful" threshold.

\textbf{(i)} By a (second-order) Taylor approximation around $f(\w_k)$ and applying the Cauchy-Schwarz inequality, we have:
\begin{equation}
\begin{aligned}
f(\w_k+\s_k)-m(\s_k)=&(\nabla f(\w)-\g_k)^\intercal \s_k + \frac{1}{2}\s_k^\intercal ((\Hm(\w_k+\tau\s_k)-\Bm_k)\s_k-\frac{\sigma_k}{3}\norm{\s}^3\\
\leq & \norm{\evec_k}\norm{\s_k}+\frac{1}{2}\norm{((\Hm(\w_k+\tau\s_k)-\Bm_k)\s_k}\norm{\s_k},
\end{aligned}
\end{equation}
where the term $\norm{\evec_k}\norm{\s_k}$ is extra compared to the case of deterministic gradients.

\textbf{(ii)} Regarding the second part we note that if $\s_k$ satisfies (\ref{eq:3.11}), we have by the definition of $R_k$ and equation (\ref{temp:R}) that
\begin{equation}
\begin{aligned}
f(\w_k)-m_k(\s_k) =& \frac{1}{2}\s_k^\intercal B\s_k + \frac{2}{3}\sigma_k\norm{\s_k}^3\\
\geq & \frac{1}{2} R_{\min} \norm{\s_k}^2,
\end{aligned}
\end{equation}
which negated gives the desired bound on (ii). All together, the upper bound on $r_k$ is written as
\begin{equation}\label{eq:bound_on_r}
r_k\leq \frac{1}{2} \norm{\s_k}^2 \left(\frac{2\norm{\evec_k}}{\norm{\s_k}}+\frac{\norm{((\Hm(\w_k+\tau \s_k)-\Bm_k)\s_k}}{\norm{\s_k}}-(1-\eta_2)R_{\min}\right).
\end{equation}
Let us add and subtract $\Hm(\w_k)$ to the second summand and apply the triangle inequality

\begin{equation}\label{eq:bound_on_r}
r_k\leq \frac{1}{2} \norm{\s_k}^2 \left(\frac{2\norm{\evec_k}}{\norm{\s_k}}+\frac{\norm{(\Hm(\w_k+\tau \s_k)-\Hm_k)\s_k}+\norm{(\Hm_k-\Bm_k)\s_k}}{\norm{\s_k}}-(1-\eta_2)R_{\min}\right).
\end{equation}

Now applying $\norm{\Am \supervector}\leq \norm{\Am}\norm{\supervector}$ we get

\begin{equation}\label{eq:bound_on_r}
r_k\leq \frac{1}{2} \norm{\s_k}^2 \left(\frac{2\norm{\evec_k}}{\norm{\s_k}}+\norm{\Hm(\w_k+\tau \s_k)-\Hm_k}+\norm{(\Hm_k-\Bm_k)}-(1-\eta_2)R_{\min}\right).
\end{equation}

We have already established in Lemma \ref{l:eventu_full_samples} that $\norm{\evec_k}\rightarrow 0$ and $\norm{(\Hm_k-\Bm_k)}\rightarrow 0$. Together with Lemma \ref{l:upper_bound_on_s} and the assumption $\norm{\nabla f_k}\rightarrow 0$ this implies $\|\s_k\|\rightarrow0$. Furthermore, since $\tau\in [0,1]$ we have that $\norm{\w_k+\tau\s_k}\leq \norm{\w_k+\s_k}\leq \norm{\s_k}$. Hence, $\Hm(\w_k+\tau\s_k)$ and $\Hm(\w_k)$ eventually agree.  Finally, $\eta_2<1$ and $R_{\min}>0$ such that $r_k$ is negative for all $k$ sufficiently large, which implies that every such iteration is very successful.
\begin{flushright}
$\square$
\end{flushright}

\textbf{\textit{Proof of Theorem \ref{t:quadratic_convergence_exp}:}}

From Lemma \ref{l:bounds_on_sigma} we have $\sigma_k\leq \sigma_{sup}$. Furthermore, all assumptions needed for the step size bounds of Lemma \ref{l:lower_bound_on_s} and \ref{l:upper_bound_on_s} hold. Finally, Lemma \ref{l:eventu_succ_iter} gives that all iterations are eventually successful. Thus, we can combine the upper (\ref{eq:upper_bound_on_s}) and lower (\ref{eq:better_lower_bound_on_s}) bound on the stepsize for all $k$ sufficiently large to obtain
\begin{equation}
\frac{1}{R_{\min}}(\norm{\nabla f(\w_k)} + \norm{\evec_k}) \geq \norm{\s_k} \geq \kappa_s\sqrt{\norm{\nabla f(\w_{k+1})}}
\end{equation}
which we can solve for the gradient norm ratio
\begin{equation}\label{temp:quadratic}
\dfrac{\norm{\nabla f(\w_{k+1})}}{\norm{\nabla f(\w_k)}^2} \leq \left(\dfrac{1}{R_{\min}\kappa_s}\left(1+\frac{\norm{\evec_k}}{\norm{\nabla f(\w_k)}}\right)\right)^2.
\end{equation}
Consequently, as long as the right hand side of (\ref{temp:quadratic}) stays below infinity, i.e. $\norm{\evec_k}/\norm{\nabla f(\w_k)} \not\rightarrow \infty$, we have quadratic convergence of the gradient norms. From Lemma \ref{l:eventu_full_samples} we have that $\norm{\evec_k}\rightarrow 0$ as $k \rightarrow \infty$ w.h.p. and furthermore $\kappa_s$ is bounded above by a constant and $R_{\min}$ is a positive constant itself which gives quadratic convergence of the gradient norm ratio with high probability. Finally, the convergence rate of the iterates follows from a Taylor expansion around $\g_k$.
\begin{flushright}
$\square$
\end{flushright}

\subsubsection{First order global convergence}
Note that the preliminary results Lemma \ref{l:lower_bound_function_decrease} and \ref{l:lower_bound_on_s} allow us to lower bound the function decrease of a successful step in terms of the \textit{full} gradient $\nabla f_{k+1}$. Combined with Lemma \ref{l:bounds_on_sigma}, this enables us to give a \textit{deterministic} global convergence guarantee while using only \textit{stochastic} first order information\footnote{Note that this result can also be proven without Lipschitz continuity of $H$ and less strong agreement conditions as done in Corollary 2.6 in \citep{cartis2011adaptive}.}.

\textbf{\textit{Proof of Theorem \ref{t:1st_order_guarantee}:}}

We will consider two cases regarding the number of successful steps for this proof.

Case (i): \methodname takes only finitely many successful steps. Hence, we have some index $k_0$ which yields the very last successful iteration and all further iterates stay at the same point $\x_{k_0+1}$. That is $\x_{k_0+1}=\x_{k_0+i},\:\forall\: i\geq1$. Let us assume that $\norm{\nabla f(\x_{k_0+1})}=\epsilon>0$, then 
\begin{equation}
\norm{\nabla f(\x_{k})}=\epsilon ,\: \forall \: k\geq k_0+1 .
\end{equation}
Since, furthermore, all iterations $k\geq k_0+1$ are unsuccessful $\sigma_k$ increases by $\gamma$, such that
\begin{equation}
\sigma_k \rightarrow \infty \text{ as } k \rightarrow \infty.
\end{equation}
However, this is in contradiction with Lemma \ref{l:bounds_on_sigma}, which states that $\sigma_k$ is bounded above. Hence, the above assumption cannot hold and we have $\norm{\nabla f(\x_{k_0+1})}=\norm{\nabla f(\x^*)}=0$.

Case (ii): sARC takes infinitely many successful steps. While unsuccessful steps keep $f(\x_k)$ constant, (very) successful steps strictly decrease $f(\x_k)$ and thus the sequence $\{f(\x_k)\}$ is monotonically decreasing. Furthermore, it is bounded below per assumption and thus the objective values converge 
\begin{equation}\label{eq:f_to_0}
f(\x_k) \rightarrow f_{\inf}, \text{ as } k\rightarrow \infty.
\end{equation}
All requirements of Lemma \ref{l:lower_bound_function_decrease} and Lemma \ref{l:lower_bound_on_s} hold and we thus can use the sufficient function decrease equation (\ref{eq:sufficient_function_decrease}) to write
\begin{equation}
f(\x_k)-f_{\inf} \geq f(\x_k)-f(\x_{k+1}) \geq  \frac{1}{6}\eta_1\sigma_{\inf}\kappa_s^3\:\norm{\nabla f(\x_{k+1})}^{3/2}.
\end{equation}
Since $(f(\x_k)-f_{\inf}) \rightarrow 0 \text{ as } k\rightarrow \infty$, $\sigma_{\inf} >0, \eta_1>0$ and $\kappa_s^3 \geq 0$ (as $\sigma_{\sup} <\infty$), we must have $\norm{\nabla f(\x_k)}\rightarrow 0$, giving the result.
\begin{flushright}
$\square$
\end{flushright}

\subsubsection{Second order global convergence and worst case iteration complexity}
For the proofs of Theorem \ref{t:2nd_order_guarantee} and Theorem \ref{t:fowcc} we refer the reader to Theorem 5.4 in \citep{cartis2011adaptive} and Corollary 5.3 in \citep{cartis2011adaptive2}. Note that, as already laid out above in the proofs of Lemma \ref{l:bounds_on_sigma} and Lemma \ref{l:lower_bound_function_decrease}, the constants involved in the convergence Theorems change due to the stochastic gradients used in our framework.


\subsection{Details concerning experimental section}\label{sec:exp_details}

We here provide additional results and briefly describe the baseline algorithms used in the experiments as well as the choice of hyper-parameters. All experiments were run on a CPU with a 2.4 GHz nominal clock rate.

\paragraph{Datasets}
The real-world datasets we use represent very common instances of Machine Learning problems and are part of the libsvm library \citep{chang2011libsvm}, except for \textit{cifar} which is from \citet{krizhevsky2009learning}. A summary of their main characteristic can be found in table 1. The multiclass datasets are both instances of so-called image classification problems. The \textit{mnist} images are greyscale and of size $28\times 28$. The original \textit{cifar} images are $32\times 32\times 3$ but we converted them to greyscale so that the problem dimensionality is comparable to \textit{mnist}. Both datasets have $10$ different classes, which multiplies the problem dimensionality of the multinomial regression by $10$.

\begin{table}[H]
\centering
\begin{tabular}{l|lllll}
                    & type                           & n          & d      & $\kappa(H*)$ & $\lambda$ \\ \hline
a9a                 & Classification              & $32,561$      & $123$    & $761.8$        & $1e^{-3}$ \\
a9a nc    & Classification              & $32,561$      & $123$    & $1,946.3$       & $1e^{-3}$ \\
covtype             & Classification              & $581,012$    & $54$     & $3\cdot10^9$ & $1e^{-3}$ \\
covtype nc & Classification              & $581,012$    & $54$     & $25,572,903.1$ & $1e^{-3}$ \\
higgs               & Classification              & $11,000,000$ & $28$     & $1,412.0$       & $1e^{-4}$ \\
higgs nc   & Classification              & $11,000,000$ & $28$     & $2,667.7$       & $1e^{-4}$ \\
mnist               & Multiclass & $60,000$     & $7,840$   &      $10,281,848$        &    $1e^{-3}$      \\
cifar               & Multiclass & $50,000$     & $10,240$ &      $1\cdot10^9$        &    $1e^{-3}$      
\end{tabular}
\label{table:datasets}
\caption{Overview over the real-world datasets used in our experiments with convex and non-convex (nc) regularizer. $\kappa(H*)$ refers to the condition number of the Hessian at the optimizer and $\lambda$ is the regularization parameter applied in the loss function and its derivatives.}

\end{table}

\begin{figure*}[h!]
	\begin{center}
          \begin{tabular}{@{}c@{\hspace{0.5mm}}c@{\hspace{0.5mm}}c@{}}
            \vspace{-5.5pt}
            \includegraphics[width=0.33\linewidth]{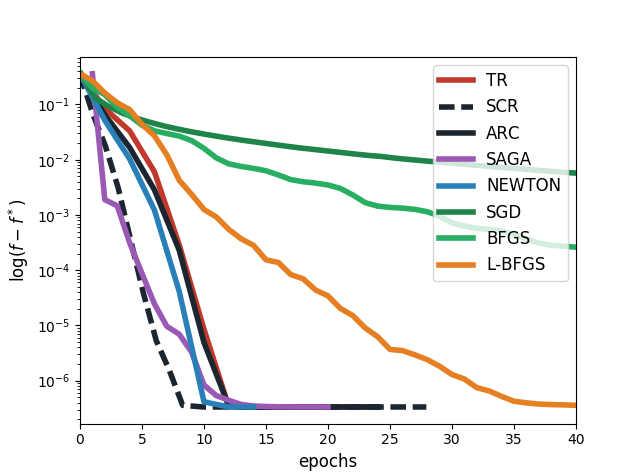} &
            \includegraphics[width=0.33\linewidth]{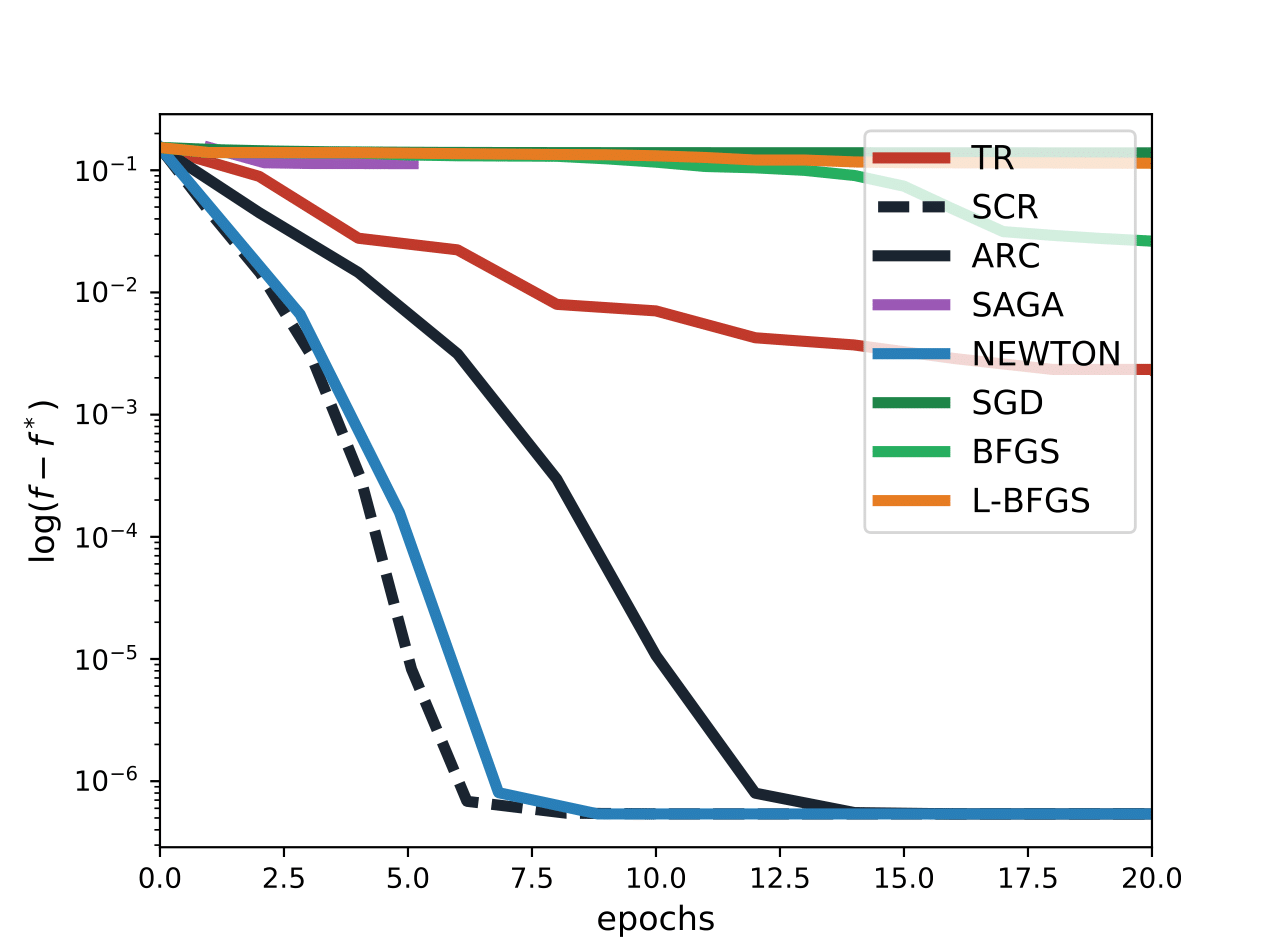}  &
             \includegraphics[width=0.33\linewidth]{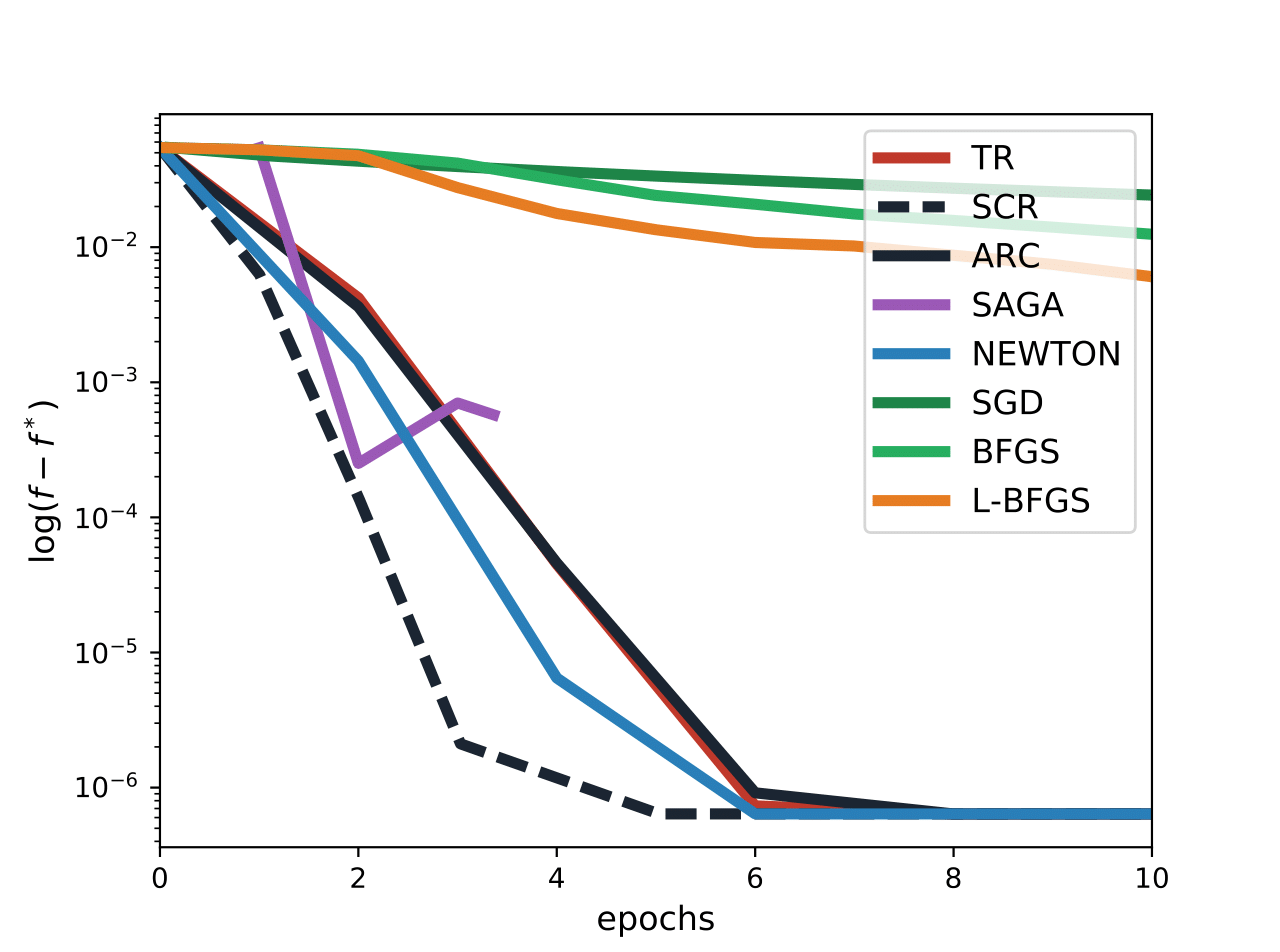}    \\ 

             \includegraphics[width=0.33\linewidth]{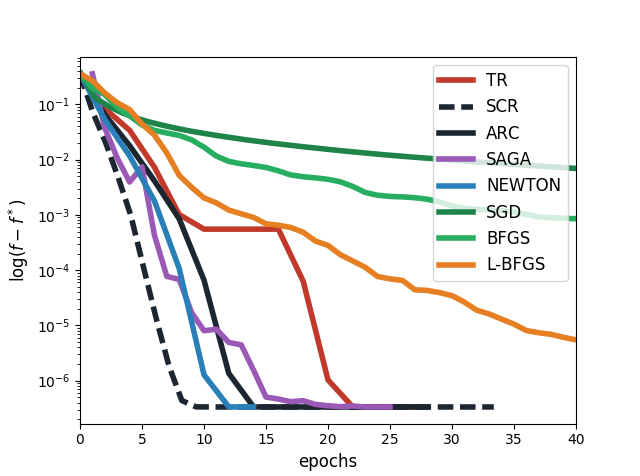} &
            \includegraphics[width=0.33\linewidth]{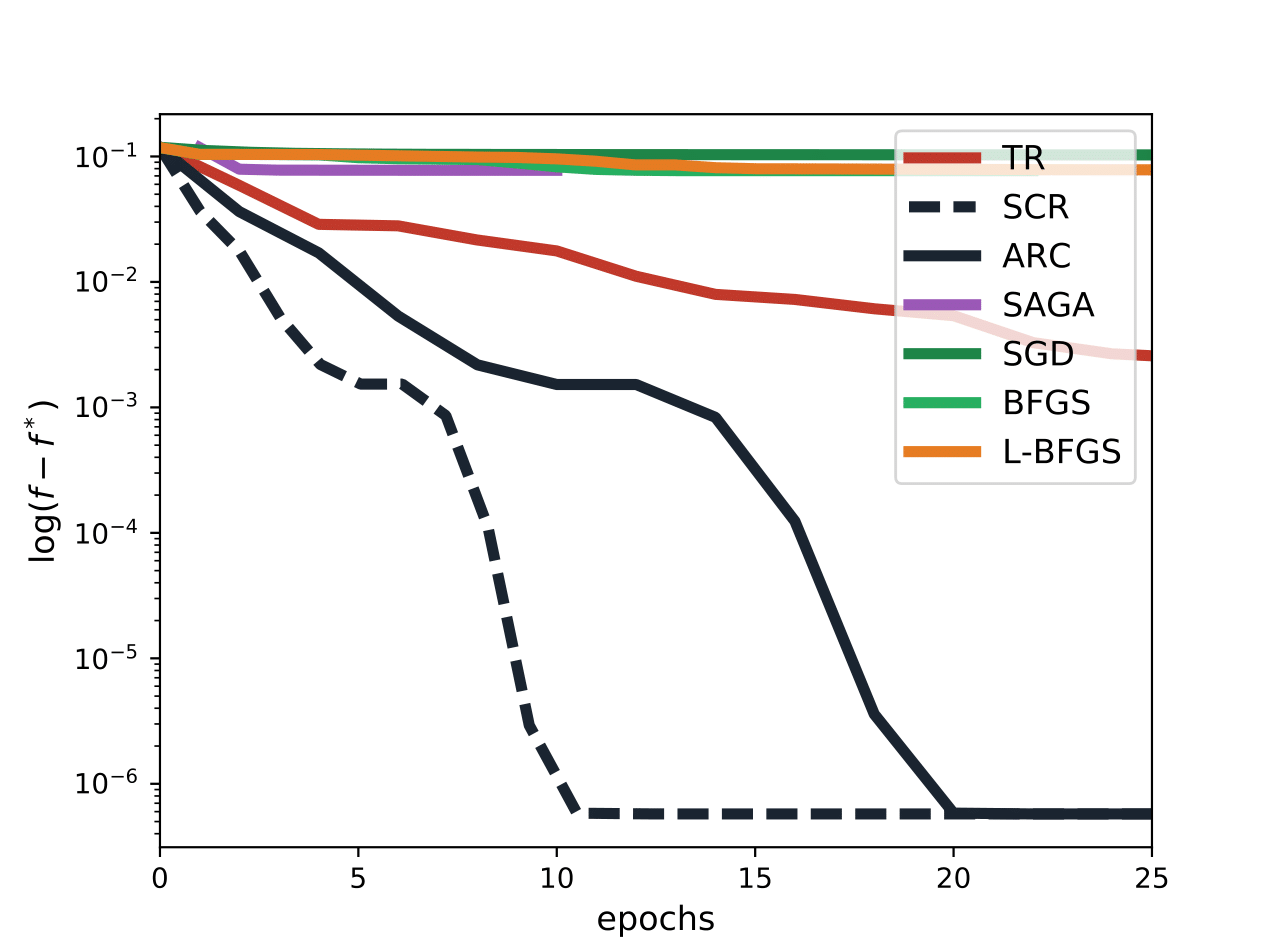}  &
              \includegraphics[width=0.33\linewidth]{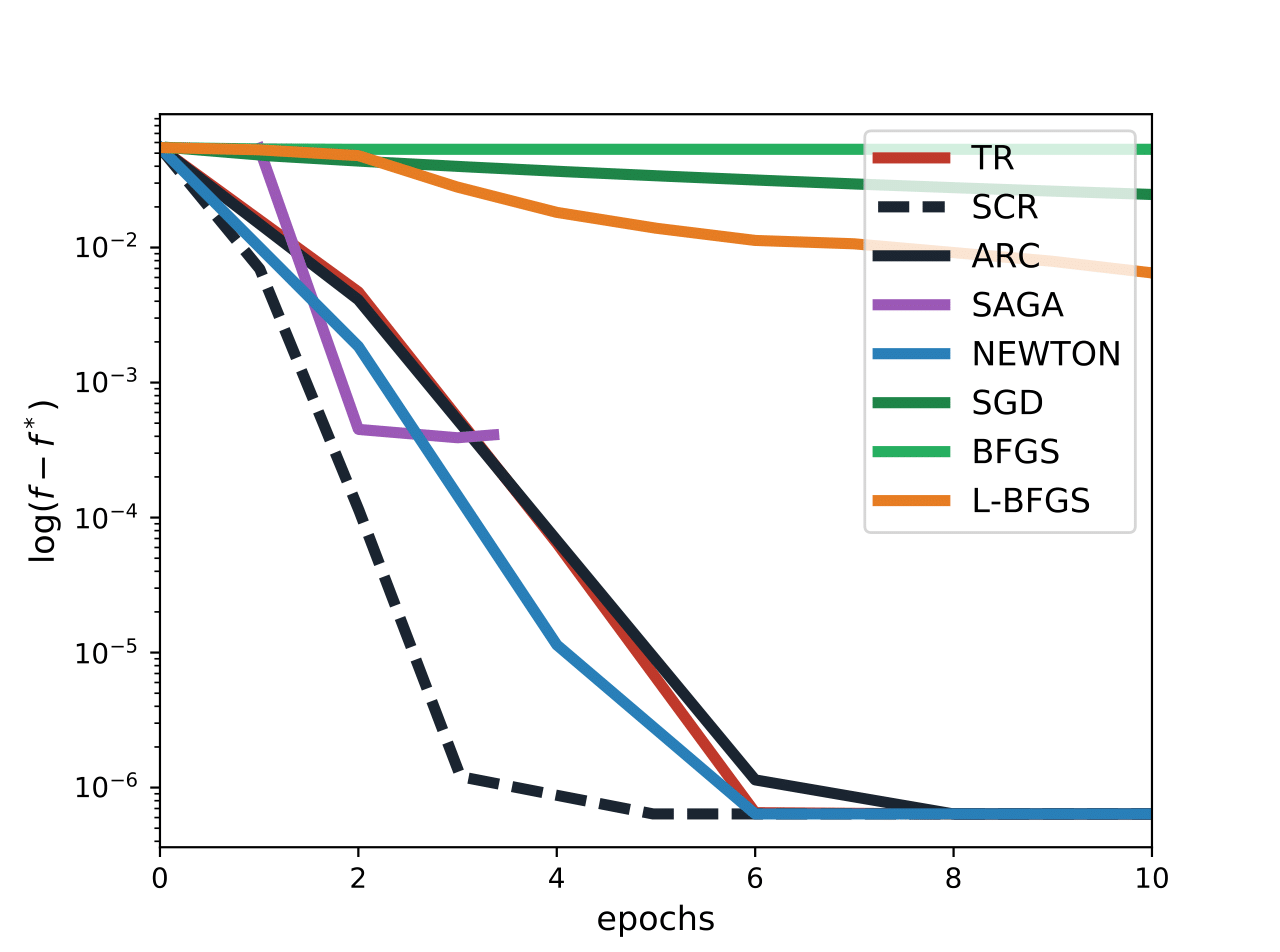}     \\ 
         	1. \footnotesize{{ A9A}} &
            2. \footnotesize{{ COVTYPE}} & 
            3. \footnotesize{{ HIGGS}} 
	  \end{tabular}
          \caption{ {\it Results from Section 5 over epochs}.~Top (bottom) row shows the log suboptimality of convex (non-convex) regularized logistic regressions over epochs (average of 10 independent runs).}
          \label{fig:results2}
	\end{center}
\end{figure*}

\paragraph{Benchmark methods}
\begin{itemize}
\item Stochastic Gradient Descent (SGD): To bring in some variation, we select a mini-batch of the size $\lceil n/10 \rceil$ on the real world classification- and $\lceil n/100 \rceil$ on the multiclass problems. On the artificial datasets we only sample $1$ datapoint per iteration and update the parameters with respect to this point. We use a problem-dependent, constant step-size as this yields faster initial convergence~\citep{hofmann2015variance},\citep{roux2012stochastic}.
\item SAGA: is a variance-reduced variant of SGD that only samples 1 datapoint per iteration and uses a constant step-size.
\item Broyden-Fletcher-Goldfarb-Shanno (BFGS) is the most popular and stable Quasi-Newton method. 
\item Limited-memory BFGS is a variant of BFGS which uses only the recent $K$ iterates and gradients to construct an approximate Hessian. We used $K=20$ in our experiments. Both methods employs a line-search technique that satisfies the strong Wolfe condition to select the step size.
\item NEWTON is the classic version of Newton's method which we apply with a backtracking line search.
\end{itemize}

For L-BFGS and BFGS we used the implementation available in the optimization library of \href{https://docs.scipy.org/doc/scipy-0.18.1/reference/optimize.html}{scipy}. All other methods are our own implementation. The code for our implementation of SCR is publicly available on the authors' webpage.

\paragraph{Initialization.} All of our experiments were started from the initial weight vector $\w_0:=(0,\ldots,0)$.
 
\paragraph{Choice of parameters for ARC and SCR.}
The regularization parameter updating is analog to the rule used in the reported experiments of \cite{cartis2011adaptive}, where $\gamma=2$. Its goal is to reduce the penalty rapidly as soon as convergence sets in, while keeping some regularization in the non asymptotic regime. A more sophisticated approach can be found in \cite{gould2012updating}. In our experiments we start with $\sigma_0=1, \eta_1 =0.2, \text{ and } \eta_2=0.8$ as well as an initial sample size of $5\%$.

\paragraph{Influence of dimensionality}
To test the influence of the dimensionality on the progress of the above applied methods we created artificial datasets of three different sizes, labeled as \textit{gaussian s}, \textit{gaussian m} and \textit{gaussian l}.
\begin{table}[h!]
\centering
\begin{tabular}{l|lllll}
                    & type                           & n          & d      & $\kappa(H^*)$ & $\lambda$ \\ \hline
gaussian s  & Classification              & $50,000$ & $100$     & $2,083.3$       & $1e^{-3}$ \\
gaussian m  & Classification              & $50,000$ & $1,000$     & $98,298.9$       & $1e^{-3}$ \\
gaussian l  & Classification              & $50,000$ & $10,000$     & $1,167,211.3$       & $1e^{-3}$ \\    
\end{tabular}
\caption{Overview over the synthetic datasets used in our experiments with convex regularizer}

\end{table}

The feature vectors $X=(\x_1,\x_2,...,\x_d), \x_i \in \mathbb{R}^n$ were drawn from a multivariate Gaussian distribution
\begin{equation}
X \sim \mathcal{N}(\mu,\,\Sigma)
\end{equation}
 with a mean of zero $\mu=(0,\ldots,0)$ and a covariance matrix that has reasonably uniformly distributed off-diagonal elements in the interval $(-1,1)$. 

\begin{figure*}[h!]
	\begin{center}
          \begin{tabular}{@{}c@{\hspace{0.5mm}}c@{\hspace{0.5mm}}c@{}}
             \vspace{-6pt}
              \includegraphics[width=0.33\linewidth]{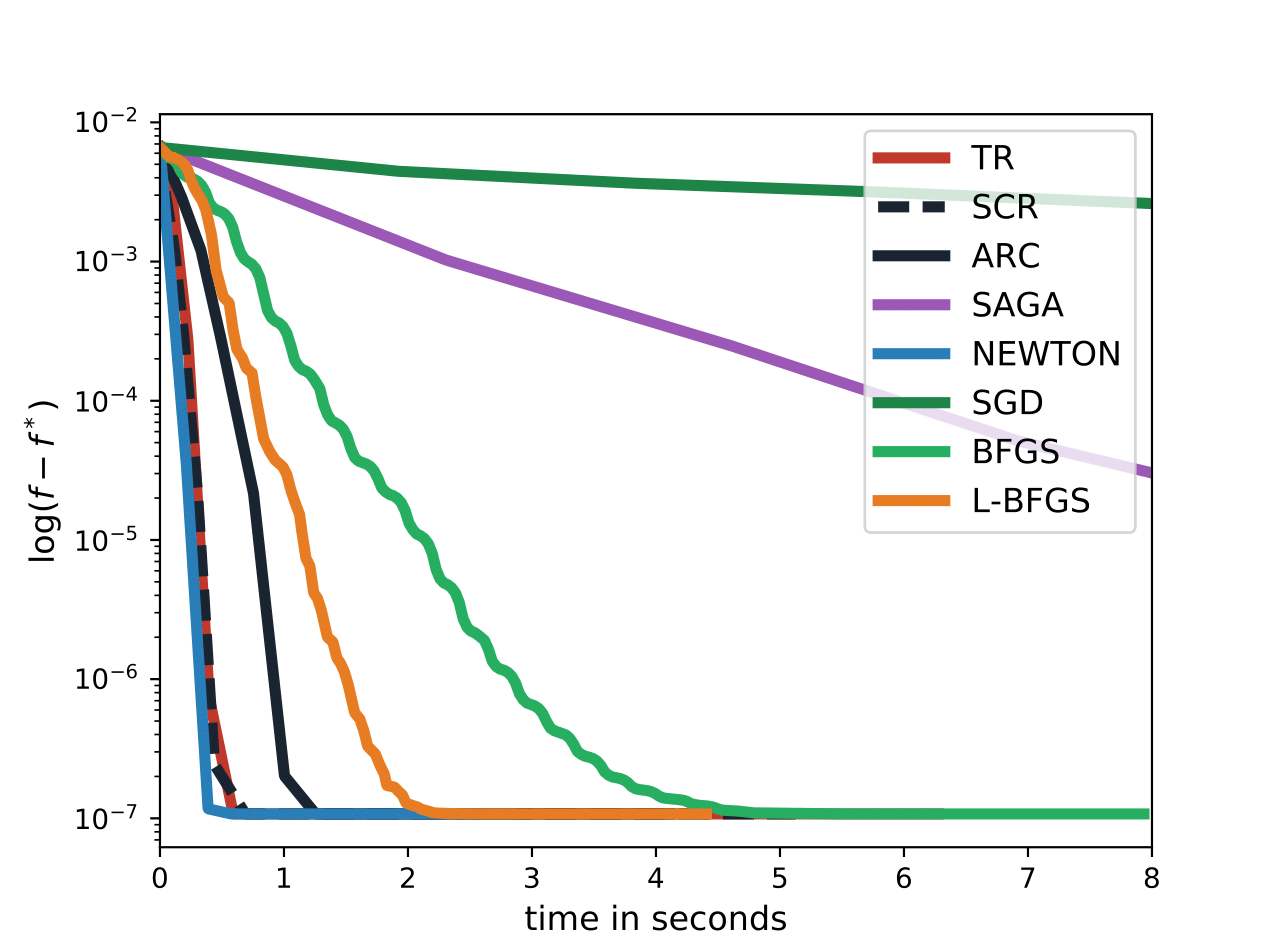} &
            \includegraphics[width=0.33\linewidth]{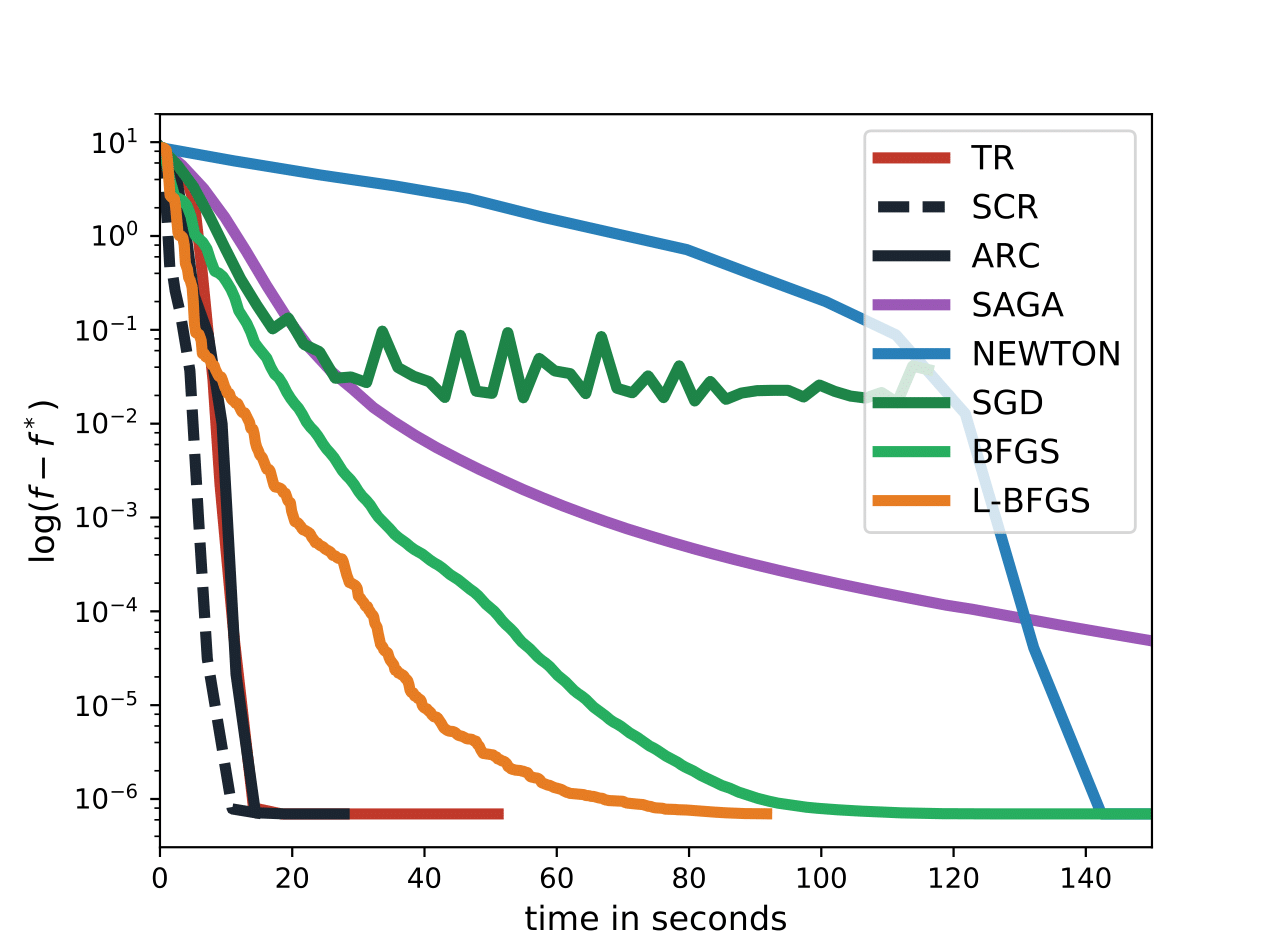} & 
             \includegraphics[width=0.33\linewidth]{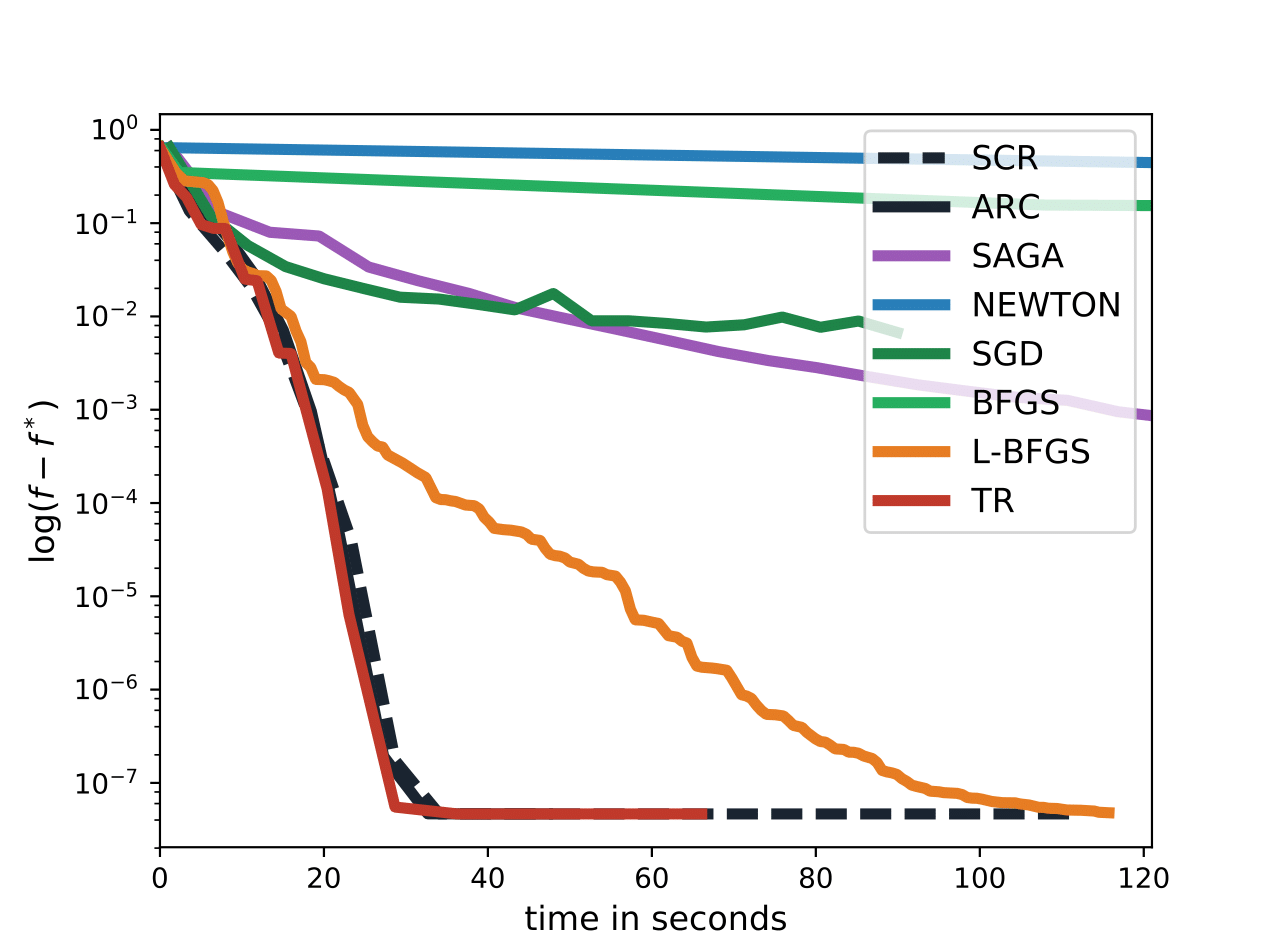}  \\ 
             \includegraphics[width=0.33\linewidth]{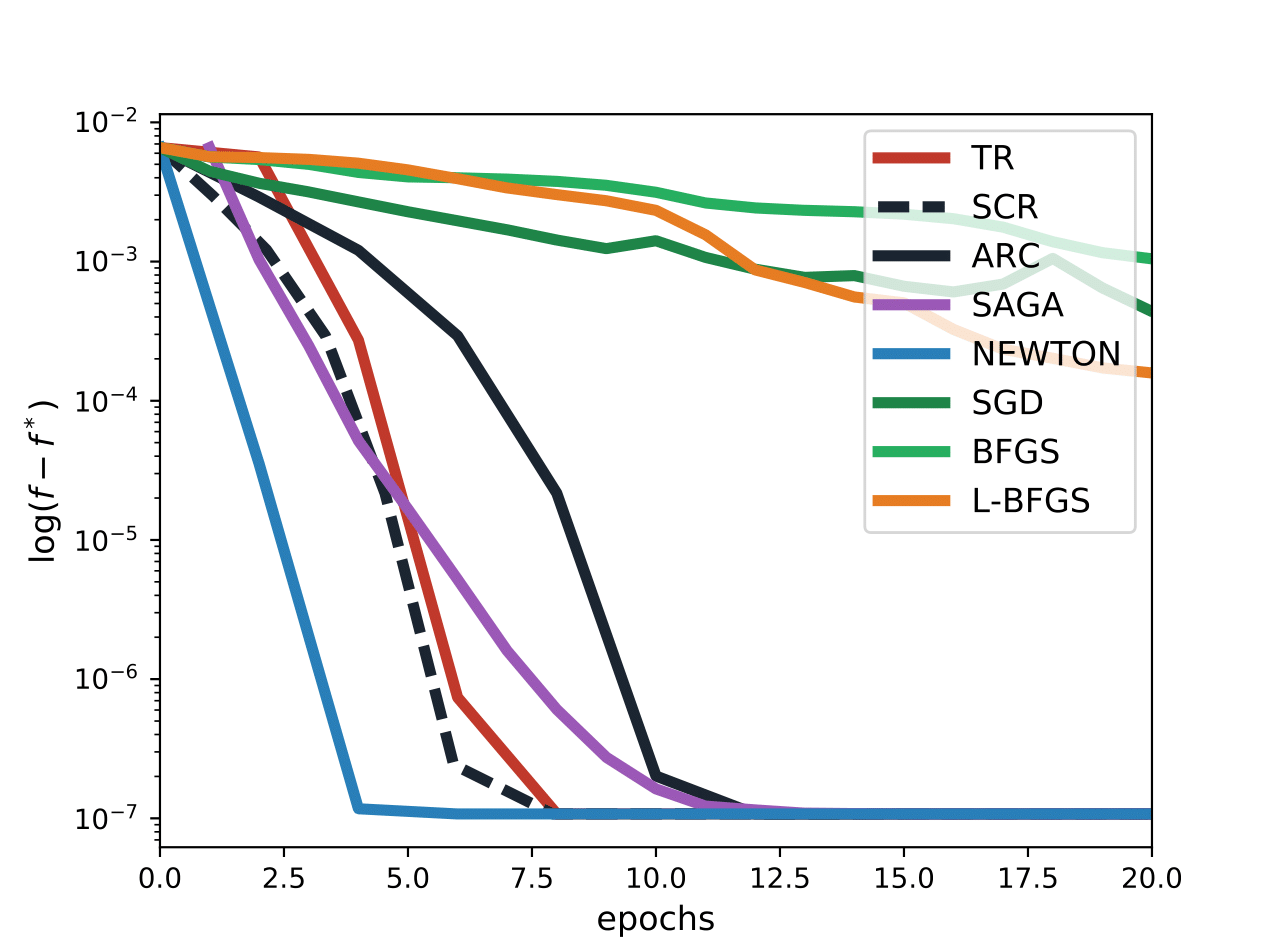} &
            \includegraphics[width=0.33\linewidth]{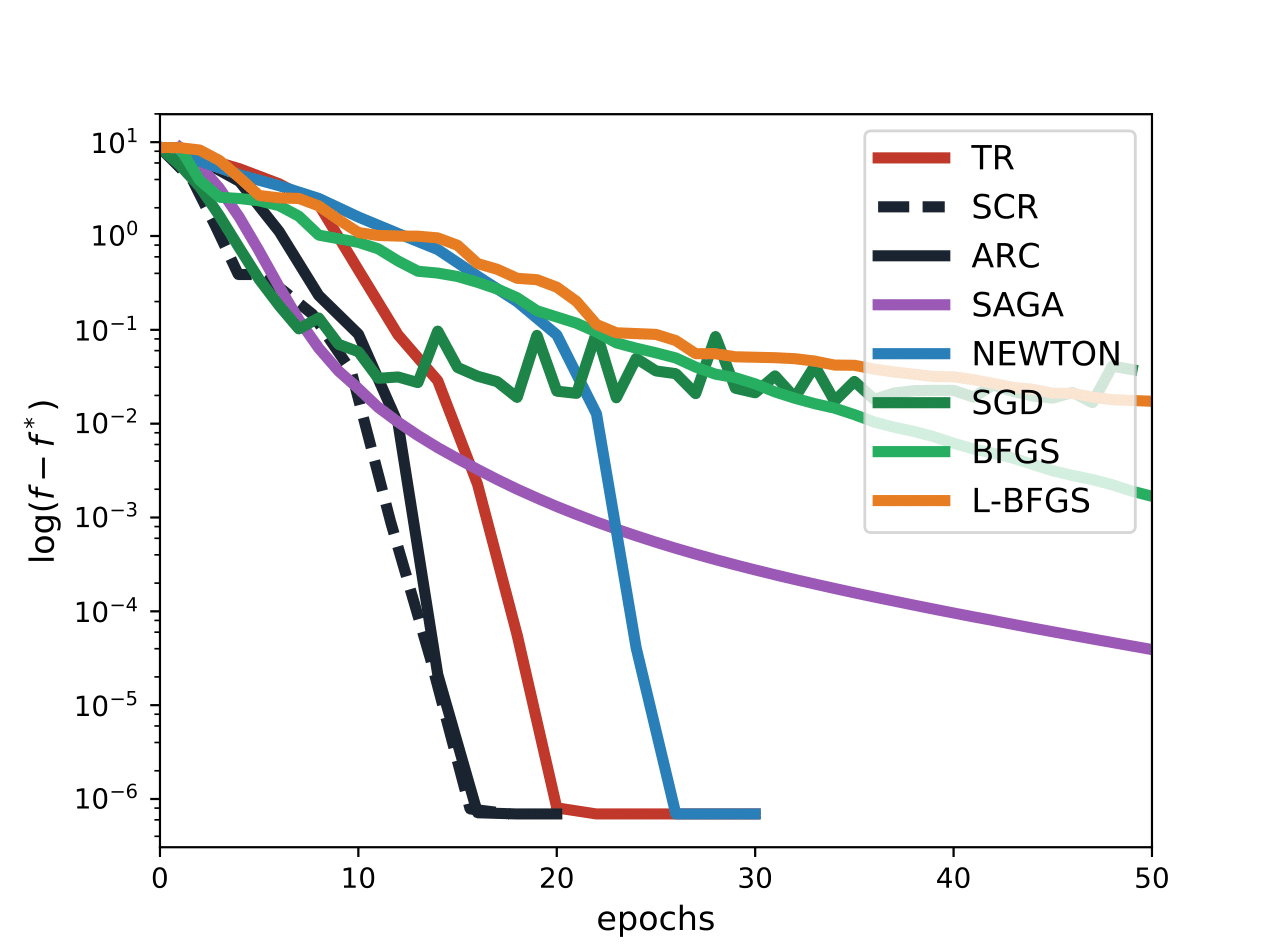}  &
              \includegraphics[width=0.33\linewidth]{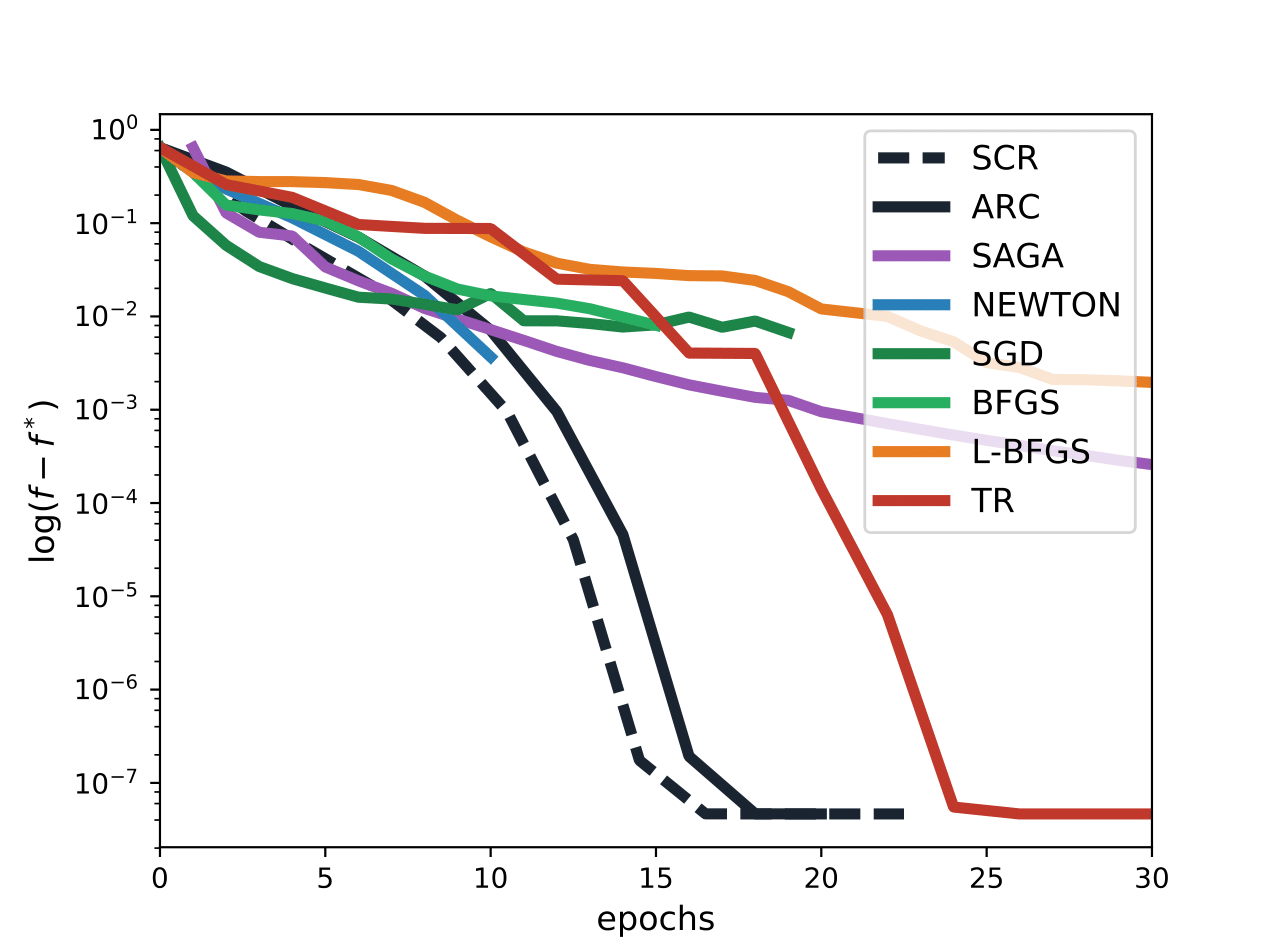}     \\ 
         	1. { GAUSSIAN S } &
            2. { GAUSSIAN M} &
            3. { GAUSSIAN L} 
	  	\end{tabular}
	  	 \caption{Top (bottom) row shows the log suboptimality of convex regularized logistic regressions over time (epochs) (average of 10 independent runs).}
          \label{fig:results3}
	\end{center}
\end{figure*}
As expected, the classic Newton methods suffers heavily from an increase in the dimension. The regularized Newton methods on the other hand scale comparably very well since they only need indirect access to the Hessian via matrix-vector
products. Evidently, these methods outperform the quasi-newton approaches even in high dimensions. Among these, the limited memory version of BFGS is significantly faster than its original variant.

\begin{figure*}[h!]
	\begin{center}
          \begin{tabular}{@{}c@{\hspace{0.5mm}}c@{\hspace{0.5mm}}c@{}}
           \vspace{-6pt}
             \includegraphics[width=0.33\linewidth]{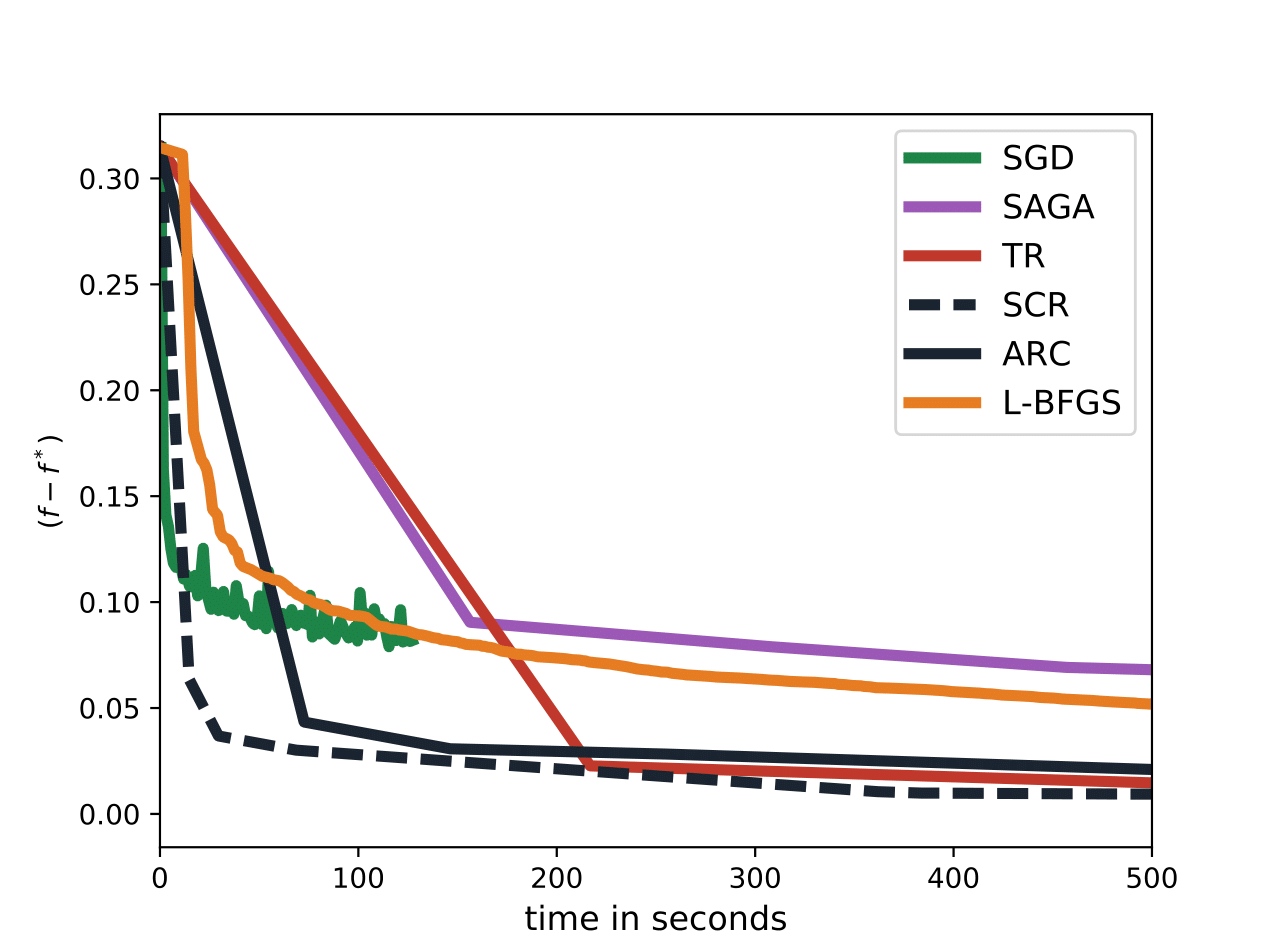} &
            \includegraphics[width=0.33\linewidth]{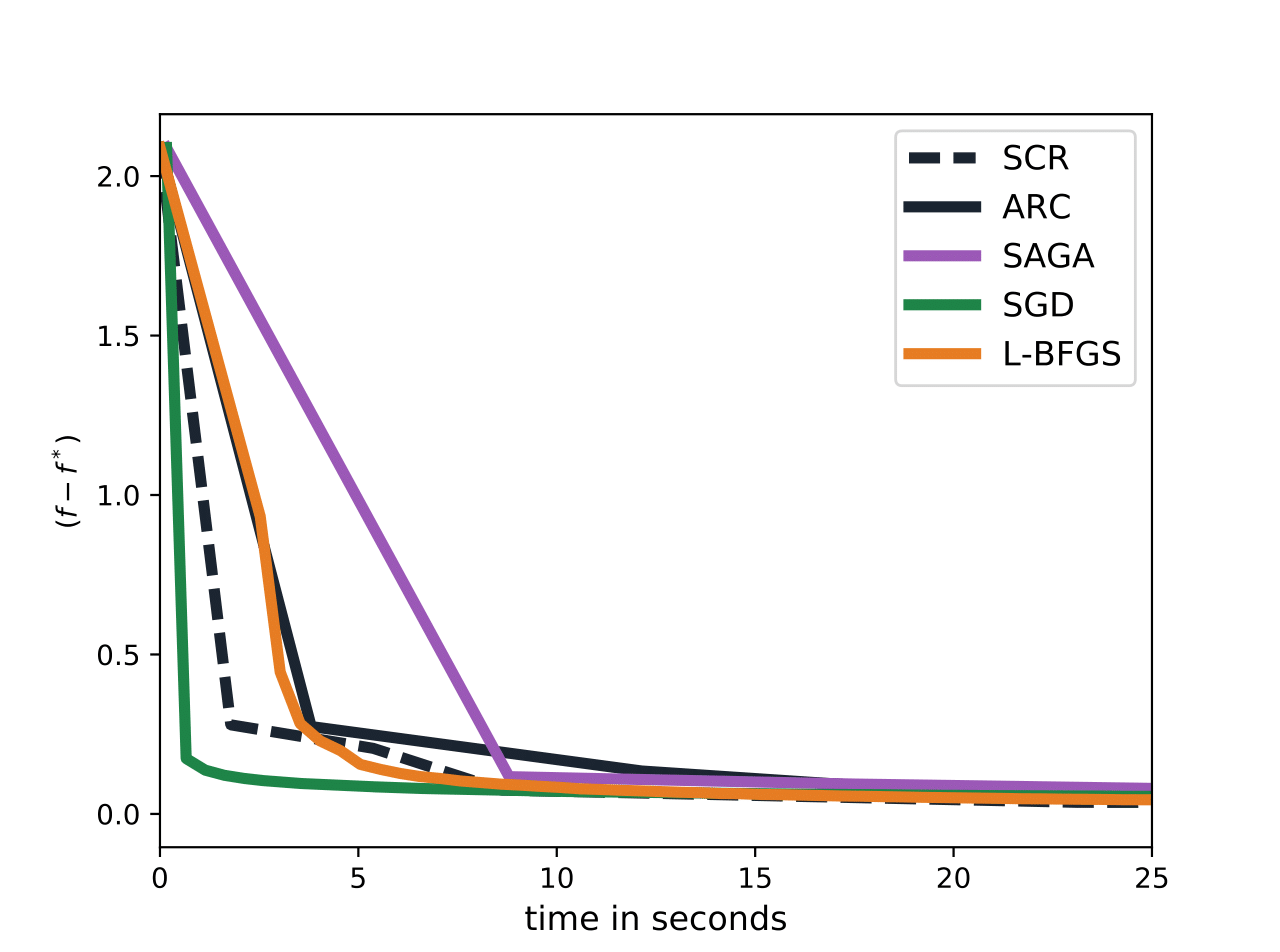} & \\
             \includegraphics[width=0.33\linewidth]{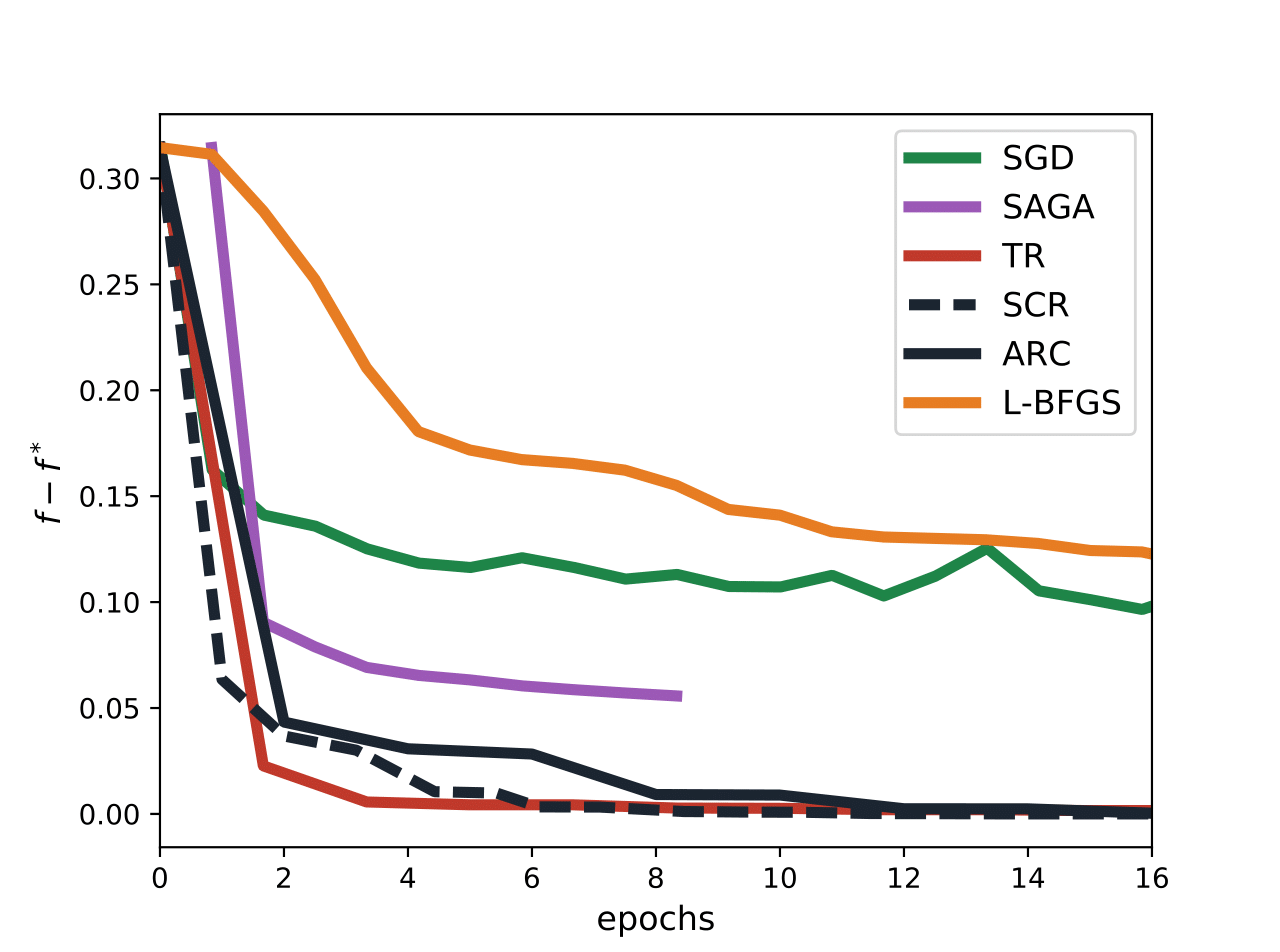} &
            \includegraphics[width=0.33\linewidth]{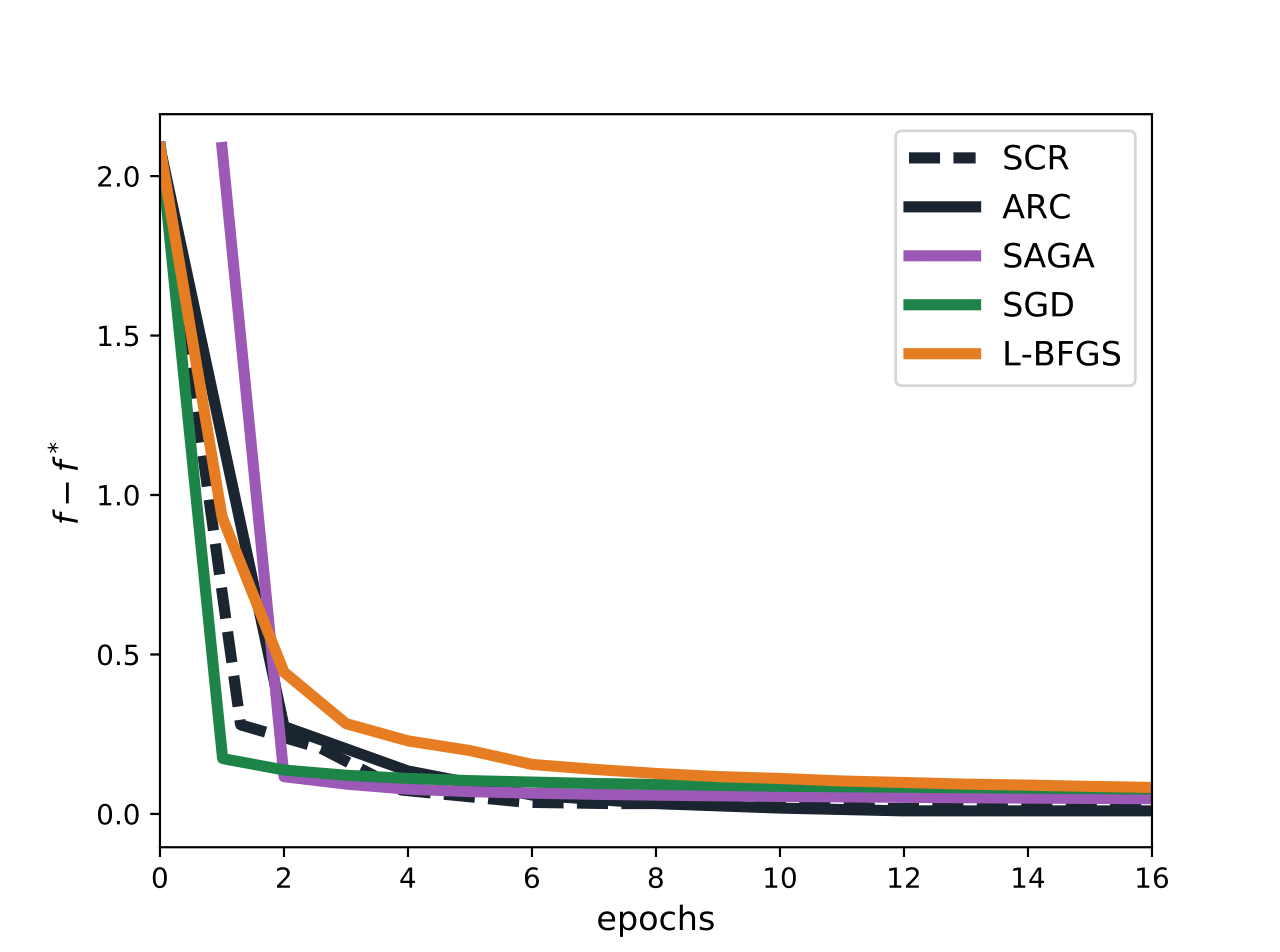}  & \\
         	1. { CIFAR} &
            2. { MNIST} 
	  	\end{tabular}
          \caption{Top (bottom) row shows suboptimality of the empirical risk of convex regularized multinominal regressions over time (epochs)}
          \label{fig:results4}
	\end{center}
\end{figure*}

\paragraph{Multiclass regression}
In this section we leave the trust region method out because our implementation is not optimized towards solving multi-class problems. We do not run  Newton's method or BFGS either as the above results suggests that they are unlikely to be competitive. Furthermore, Figure \ref{fig:results4} does not show logarithmic but linear suboptimality because optimizing these problems to high precision takes very long and yields few additional benefits. For example, the 25th SCR iteration drove the gradient norm from $3.8\cdot 10^{-5}$ to $5.6\cdot 10^{-8}$ after building up a Krylov space of dimensionality $7800$. It took 9.47 hours and did $\textit{not}$ change any of the first $13$ digits of the loss. As can be seen, SCR provides early progress at a comparable rate to other methods but gives the opportunity to solve the problem to high precision if needed.

\end{document}